\documentclass[11pt]{article}
 \pdfoutput=1
\usepackage[dvipsnames]{xcolor}
\pagecolor{white}
\usepackage[utf8x]{inputenc}
\usepackage[square, numbers, ]{natbib}
\usepackage{ifpdf} 
\usepackage{graphicx}
\usepackage{tikz}

\usepackage[utf8x]{inputenc}
\usepackage[square, numbers]{natbib}
\usepackage{enumitem}
\usepackage{amsmath, amsfonts}
\usepackage{tikz}
\usetikzlibrary{decorations.pathreplacing,calligraphy}
\usepackage{subcaption}
\usepackage{xcolor}
\usepackage{authblk}
\usepackage{ragged2e}
\usepackage{blindtext}
\usepackage{float}
\usepackage{tikz-network}
\usetikzlibrary{calc,arrows,positioning,shapes.arrows}
\usepackage{hyperref}
\usepackage{authblk}
\newcommand{\beq}{\begin{equation}}
\newcommand{\eeq}{\end{equation}}
\newcommand \R {{ \mathbb R}}
\newcommand \N {{ \mathbb N}}
\newcommand{\graph}{\mathbf{G}}
\newcommand{\graphset}{\mathcal{G}}

\usepackage{graphicx}
\usepackage{dcolumn}
\newcolumntype{d}[1]{D{.}{.}{#1}}
\newcommand{\cc}[1]{\multicolumn{1}{c}{#1}}   
\newcommand{\ccl}[1]{\multicolumn{1}{c|}{#1}}   

\newcommand{\psup}[1]{^{(#1)}}

\newcommand{\tablefontsize}{\small}

\usepackage{chngcntr}
\counterwithout{figure}{section}

\DeclareMathOperator{\enc}{Enc}
\DeclareMathOperator{\dec}{Dec}

\DeclareMathOperator{\softmax}{softmax}

\DeclareMathOperator{\sigmoid}{sigmoid}

\DeclareMathOperator{\gt}{Gt}
\DeclareMathOperator{\pre}{Pre-Proc}
\DeclareMathOperator{\messagep}{Message-Passing}
\DeclareMathOperator{\post}{Post-Proc}

\begin{document}



\title{Introduction to Graph Neural Networks for Machine Learning Engineers}

\author[1,2]{James H. Tanis\thanks{Corresponding author: \texttt{jhtanis22@gmail.com}}}
\author[1]{Chris Giannella}
\author[1]{Adrian V. Mariano}
\author[2]{Daoud~Meerzaman}

\affil[1]{The MITRE Corporation, USA}
\affil[2]{National Cancer Institute, USA}
\date{}
\maketitle


\begin{abstract}
Graph neural networks are deep neural networks designed for graphs with attributes attached to nodes or edges.  
The number of research papers in the literature concerning these models is growing rapidly due to their impressive performance on a broad range of tasks.  
This survey introduces graph neural networks through the encoder-decoder framework and provides examples of decoders for a range of graph analytic tasks.  
It uses theory and numerous experiments on homogeneous graphs to illustrate the behavior of graph neural networks under different training sizes and degrees of graph complexity, with an emphasis on oversmoothing and oversquashing.
\end{abstract}







\section{Introduction} 

Relationships within data are important for everyday tasks like internet search and road map navigation as well as for scientific research in fields like bioinformatics.  
Such relationships can be described using graphs with real vectors as attributes associated with the graph’s nodes or edges; 
however, traditional machine learning models operate on arrays, so they cannot directly exploit the relationships.  
This report surveys Graph Neural Networks (GNNs), which jointly learn from both edge and node feature information, and often produce more accurate models.  
These architectures have become popular due to their impressive performance on graph analysis tasks.  
Consequently, the number of research papers on GNNs is growing rapidly, and many surveys exist.  

Some surveys discuss graph neural networks in the context of broad families such as graph networks,  
graph representation learning and geometric deep learning~\cite{battaglia2018relational, hamilton2020graph, chami2020machine, bronstein2021geometric, wu2022graph}.  
Other surveys categorize GNNs by abstracting their distinguishing properties into functional relationships~\cite{wu2020comprehensive, zhou2020graph, zhang2020deep, chami2020machine, zhou2022graph}.  
Although useful for organizational purposes, generality and abstraction can be difficult to understand for those new to the field.  
Other surveys have a narrow focus, for example to discuss efforts to improve a specific weakness in GNN architectures~\cite{rusch2023survey}, or to survey GNN work on a particular task, such as fake news detection or product recommendation~\cite{phan2023fake, gao2023survey, shao2024distributed}.  
While valuable for those interested in the task, they provide little background in GNNs and therefore assume the reader already has that knowledge.  

For this reason, a concrete introduction to GNNs is missing, so we
provide one here by combining theory with examples of their behavior in practice.    
We begin by introducing GNNs as encoder-decoder architectures. 
To provide perspective on the ways GNNs are used, we discuss common GNN applications along with examples of task-specific decoders for turning features into predictions.  
We think that studying a few important examples of GNNs well will help the reader develop a feeling for the subject that would be difficult to achieve otherwise.  
We therefore focus on three convolutional and attentional networks, GCN, GraphSAGE, and GATv2, which are commonly used both as benchmarks and as components in other GNN architectures.  
We conduct numerous experiments with these GNNs at two training sizes and on thirteen datasets of both high and low complexity.  
Transformer-based graph models are also an increasingly active direction, but they are outside the scope of this introductory treatment.

Our theoretical part also introduces five key challenges in the field—low homophily, label scarcity, oversmoothing, oversquashing, and explainability. 
In the experiments section, we focus on the first four, examining how they manifest in practice and how far they can be mitigated by simple means.  
Our experiments section has several goals: 
\begin{itemize}[noitemsep, nolistsep]
\item Compare benchmark GNNs with other graph models. 
\item Characterize how homophily, label scarcity, oversmoothing, and oversquashing affect node classification performance. 
\item Assess the extent to which these challenges can be mitigated by simple hyperparameter and architectural tuning. 
\item Contrast link prediction and node classification, highlighting differences in how they respond to depth and design choices.  
\end{itemize} 
We hope these experiments, combined with the theoretical sections, will enable newcomers to use GNNs more effectively and to improve GNN performance on their own problems.  
We also hope that experts will gain new insights from our empirical analysis.

\section{Common applications}\label{sect:graph_tasks} 
Graph neural networks are suited to a variety of graph tasks. 

\begin{enumerate} 

\item Node classification 

This task focuses on assigning labels to the nodes of a graph. 
In social networks, for example, node classification can be used to infer user roles or interests, or to predict whether a person belongs to a particular group~\cite{bhagat2011node,ahmad2019detection}. 
Beyond social networks, node classification is also used to categorize documents, videos, or webpages into different topics or classes~\cite{kipf2016semi,perozzi2014deepwalk}.

In cancer research, node classification is particularly impactful. 
Here, nodes may represent entities such as genes, proteins, regulatory elements, or clinical variables, while edges capture relationships like co-expression, signaling interactions, or the regulation of gene activity. 
By classifying these nodes, graph algorithms can help identify genes that drive cancer growth, reveal dysregulated pathways, distinguish cancer subtypes, and predict therapeutic vulnerabilities, thus advancing the goals of precision oncology~\cite{liu2023mrngcn, zhang2023hgdc, burkhart2023biology, li2024multimodal}.

\item Link prediction 

In most cases, link prediction is formulated as a binary classification problem: given two nodes, the task is to predict whether an edge between them should exist. 
In social networks, for example, link prediction is used to infer current or future relationships between individuals~\cite{pandey2019comprehensive}. 
In recommender systems, it can be viewed as predicting edges in a bipartite graph between users and items, where an edge indicates that a user is likely to interact with or prefer a product~\cite{koren2009matrix,wu2020graph}. 
Link prediction also includes discovering missing relations in knowledge graphs, a task commonly referred to as knowledge graph completion~\cite{nickel2015review,arora2020survey}.

In bioinformatics, link prediction is often used to infer missing or previously unknown relationships between biological entities. 
For example, models can suggest new gene--gene or protein--protein interactions, identify possible control links between genes that act as switches and the genes they switch on or off~\cite{chen2022genelink}, and uncover potential drug--target pairs~\cite{yu2022hgdti,shao2022dtiheta}. 
In cancer research, link prediction can help find important gene interactions that drive tumor growth~\cite{chen2022genelink}, discover pairs of genes where blocking both is especially harmful to cancer cells but less damaging to healthy cells~\cite{wang2021kg4sl,liu2022pilsl}, and suggest new connections between drugs and specific cancer types or diseases~\cite{abbas2021druglink}, potentially supporting biomarker discovery and the development of new therapies.

\item Community detection 

Community detection algorithms group nodes into clusters based on some notion of similarity or connectivity that depends on the problem setting. 
Most classical methods are not machine-learning based~\cite{smith2020guide,yang2016comparative}, although more recent approaches can be trained in an unsupervised or semi-supervised way~\cite{bandyopadhyay2021unsupervised,jin2019graph}. 
Applications include identifying social groups in social networks~\cite{wang2020hierarchical}, entity resolution~\cite{tauer2019incremental}, fraud detection~\cite{maddila2020crime}, text clustering (e.g., grouping Reddit posts by topic)~\cite{hamilton2017inductive}, and graph visualization~\cite{wongsuphasawat2017visualizing,burch2017visualizing}.

\item Node regression and edge regression 

Traffic prediction aims to forecast conditions such as speed and volume in the near future based on road sensors, supporting tasks like travel time estimation and route recommendation~\cite{yin2020comprehensive,derrow2021eta}. 
In this setting, the road network is modeled as a graph with intersections as nodes and road segments as edges, and sensors can be treated as additional nodes attached to this network. 
Predicting continuous traffic variables at sensor locations is therefore a node regression problem. 
Less commonly, edge regression is used to predict weights on road segments that represent traffic flow or counts~\cite{schaub2018flow}. 
Other examples of node regression include predicting house prices and weather variables~\cite{klemmer2021positional}, or forecasting the amount of internet traffic to web pages~\cite{rozemberczki2021multi}.

\item Graph classification and graph regression 

Traditionally, determining a molecule’s properties requires time-consuming and expensive laboratory experiments. 
Molecular property prediction is therefore a key task in materials science and drug discovery, where the goal is to estimate properties quickly and cheaply in order to guide the design of new materials and therapeutics. 
Molecules can be naturally represented as graphs, with atoms as nodes and chemical bonds as edges, and GNNs—designed to operate directly on graphs—have quickly proven well suited to this problem~\cite{reiser2022graph,fung2021benchmarking,xu2018how,pmlr-v70-gilmer17a}.
\end{enumerate}

\section{Introduction to encoder-decoder models}\label{sect:encode_decode}
~ \\ 
Throughout the paper, vectors such as $x_i$ are column vectors, and matrices such as $X = \left(x_i^T\right)_{i \in \mathcal N}$ are composed of row vectors.  

An \emph{attributed graph}, $\graph$, has a set of nodes, $\mathcal N$, as well as edges that define how the nodes relate to each other.  
We restrict our attention to undirected graphs, so the edges can be represented by a symmetric adjacency matrix, $A=(A_{ij})$ where $i,j \in \mathcal N$.   
An entry $A_{ij}$ is one if an edge connects node $i$ to node $j$
and zero otherwise.
Because the graph is \emph{attributed,} each node, $i\in \mathcal N$, has an attribute, $x_i \in \R^\ell$ for some $\ell \in \N$.  
An attributed graph is thus defined by 
\beq   \label{graph_definition}
   \graph = \left(\mathcal N, A, \left(x_i^T\right)_{i\in\mathcal N}\right).
\eeq
Encoder-decoder models on graphs are a class of machine learning models.   
Machine learning on graphs presents challenges that do not arise in conventional machine learning on vectors, because graphs are irregular data structures and do not have a natural coordinate system.  
In particular, standard convolutional neural networks for image arrays do not work on graphs, because the $k$-hop neighborhoods may be different for every node.  
Nonetheless, a typical first step for machine learning on graphs is to obtain for every node, a low-dimensional feature vector that contains all the information needed to complete the desired task. 
These feature vectors are real vectors that often contain the information needed to represent the local edge structure about each node.    

A feature vector of a node is also called a node embedding or a node representation, and collectively the feature vectors can be used for tasks on nodes, tasks on edges, or tasks on the entire graph.  
At the graph level, applying principal component analysis (PCA) or t-distributed stochastic neighbor embedding (t-SNE) to node embeddings can produce lower dimensional representations that enable visualizations to help understand how the algorithms are performing~\cite{van2008visualizing}.  
In addition, community detection algorithms use node embeddings to define the communities, either in an end-to-end fashion~\cite{Tsitsulin2020Graph} or as part of a two-step process by applying the $k$-means algorithm to the node embeddings~\cite{ng2002spectral, chunaev2020community}.  
Node embeddings also support graph classification, where in the simplest case, a mean activation over all node embeddings of the graph determines the graph class.  
More sophisticated approaches are described in~\cite{pham2017graph, zhang2018end}.  
Not surprisingly, node features are also used for node level tasks like node classification and regression~\cite{hamilton2017representation, kipf2016semi}, as well as for edge level tasks like link prediction~\cite{zhang2018link}, edge classification~\cite{kim2019edge, kipf2018neural} or edge regression~\cite{li2021cellular}.  
		 
Due to the importance of node embeddings, there are many techniques to obtain them for a range of goals and data conditions.  
Perhaps the simplest example of a node embedding is given by the rows of an adjacency matrix.  
The map $i \to (A_{i j})_{j \in  \mathcal N}$ defines node embeddings in $\R^{\vert \mathcal N \vert}$.  
These row vectors are poor feature vectors because they do not provide any structural information beyond each node's 1-hop neighborhood nor do they account for any node attributes.    

Instead, researchers may use rule-based descriptions of nodes, like centrality or clustering measurements, to produce low dimensional node representations that are more information dense, 
which may subsequently be applied to a downstream task with a traditional machine learning algorithm.  
The disadvantage of this approach is that the features are not part of the algorithm's training process, so the features are not fine-tuned to minimize the loss function.  
To do this, researchers use an encoder-decoder approach.

\subsection{Encoder-decoder framework} 

\begin{figure}
\hspace*{-0.52cm} 
\begin{tikzpicture}
\node[] at (1.55,0.2) {Encoder};
\node[rectangle,thick, draw = black, text = black, minimum size=0.3cm] (0) at  (0, 0) {\small Graph};
\node[rectangle,thick, draw = black, text = black, minimum size=0.3cm] (1) at  (4, 0)  {\small Node embeddings};
\node[] at (6.38,0.2) {Decoder};
\node[rectangle,thick, draw = black, text = black, minimum size=0.3cm, text width=1.6cm] (2) at  (8.25, 0)    {\small Model Prediction};
\node[rectangle,thick, draw = black, text = black, minimum size=0.3cm, text width=2.17cm] (3) at (8.25, 1.7)   {\small Ground Truth Function};
\node[rectangle,thick, draw = black, text = black, minimum size=0.3cm] (4) at  (11.77, 0)   {\small Loss / Eval};

\draw[-stealth, semithick] (0) -- (1); 
\draw[-stealth, semithick] (1) -- (2);
\draw[-stealth, semithick] (2) -- (4);
\draw[-stealth, semithick] (3.east) -| (4.north); 
\end{tikzpicture} 
\caption{}\label{fig:enc-dec_pipeline}
\end{figure} 
Many machine learning models adhere to an encoder-decoder framework shown in Figure~\ref{fig:enc-dec_pipeline}.  
We restrict our attention to encoders that produce node embeddings because they are the most common and doing so makes the discussion more concrete.    

Let $\mathcal N$ be a set of nodes, and let $\mathcal G$ be a set of attributed graphs on these nodes.  
The \textit{encoder} is a function,  
\beq
  \enc: \mathcal G \to \R^{\vert\mathcal N\vert \times \ell},  \quad 
  \enc(\graph) = \left(z_i^T\right)_{i \in \mathcal N}\,,
\eeq
where $\graph$ is a graph in $\mathcal G$.  
The encoder's output is a matrix of node embeddings $z_i \in \R^\ell$, where $\ell \ll \vert \mathcal N \vert$.  
 In practice, $\enc$ is an algorithm whose computation involves the graph's edges, node attributes, or both.  
A good encoder creates node embeddings that contain all of the information about each node that is required to complete the task at hand.  

Subsequently the \textit{decoder} function  
\beq
\dec: \R^{n\times\ell} \to \R^k, \quad \dec(z_{i_1}, \ldots, z_{i_n}) = \hat y \,, 
\eeq 
converts $n$ node embeddings into a prediction, $\hat y$, where $k$ is the dimension of the prediction.    
We emphasize that the decoder generally does \emph{not} simply invert the encoder.  
It is instead a kind of interpreter that ``decodes'' abstract node embeddings into predictions in order to solve the given task.  Although decoders often include learnable parameters, they are usually simple functions whose parameters enter the computation in a relatively straightforward way, for example through a linear map or inner product. 
The majority of the model's learnable parameters are usually in the encoder.  

The \textit{ground truth function}, $\gt$, of a graph, $\graph$, 
\beq
\gt: \mathcal N^n \to \R^{k}\,, \quad \gt(i_1, \ldots, i_n) = y\,,
\eeq 
provides reference information that is known about the 
nodes of $\graph$ in the form of $k$-dimensional vectors, where the integer $n$ is the same as in the decoder function.  
For the node classification problem, $n=1$, and the ground truth is the node's class represented as a $k$-dimensional vector.
The loss function compares the ground truth with the $k$-dimensional
model predictions, and 
the training and evaluation algorithms use it to assess the quality of
those model predictions.  

There does not seem to be an accepted term in the literature that is
appropriate in all contexts.  
Hamilton et~al.~\cite{hamilton2017representation} consider the case of a relationship between two nodes and call it a ``pairwise similarity'' function.   
``Pairwise similarity'' is appropriate for link prediction, where the ground truth function may be
a map $\mathcal N \times \mathcal N \to \{0,1\}$ that says whether or
not an edge exists between two nodes.
However, ``pairwise similarity'' does not describe the case of node classification, where 
the map is from $\mathcal N$ to the set of node classes.
In all cases, the role of the ground truth in the encoder-decoder
framework is the same, so we use the same name: ground truth function.

The \textit{loss} functions and \textit{evaluation} metrics link the ground truth function and the model prediction. 
Most algorithms learn model parameters by some form of gradient decent. 
Common loss functions are cross entropy loss for classification, and $L^1$ or $L^2$ loss for regression tasks.  
Common evaluation metrics for classification include accuracy, average precision, F1, and AUROC (Area Under the receiver operator characteristic Curve); for regression tasks, common evaluation metrics are RMSE (Root Mean Square Error) and MAE (Mean Absolute Error).  

\subsection{Shallow embedding examples}\label{sect:matrix_factor} 

We now present several representative examples of models that produce embedding lookups for nodes that were seen during the training process.  
These examples will illustrate the encoder-decoder framework and at the end we will note their shortcomings, which Hamilton et\ al.~\cite{hamilton2017representation, hamilton2020graph} describe.  
This will lead us to more complicated encoder-decoder models called graph neural networks in the next section.  

For each example, the input is a fixed matrix that provides a similarity statistic between any two nodes in $\mathcal N$ such as a weighted adjacency matrix.  
The output of these algorithms is a real vector (a feature vector) for each node describing the node's neighborhood structure, and taken together, they support some downstream machine learning task.    

The Laplacian eigenmaps algorithm is an early and successful nonlinear dimensionality reduction algorithm~\cite{belkin2003laplacian}.  
Given an attributed graph, $\graph$, and a 
  parameter $t > 0$, we can define a weighted adjacency
matrix, $W$, for $\graph$ by
\beq
W_{i j} = W_{i j}(t) = \left\{
\begin{aligned} 
& \exp\left(- \frac{\Vert x_i - x_j \Vert^2}{t}\right) & \textrm{ if } A_{i j} = 1\,, \\ 
& 0 & \textrm{ otherwise}\,,
\end{aligned}
\right. 
\eeq
where $i,j\in\mathcal N$, the vectors $x_i,x_j\in R^\ell$ are the graph
attributes, and $A$ is the adjacency matrix as described
in~\eqref{graph_definition}.  
Usually the Laplacian eigenmaps algorithm uses $W$ as its input, but it can also 
also use $A$ or a $k$-nearest neighbor matrix.

The Laplacian eigenmaps algorithm can be reformulated in terms of the encoder- decoder framework~\cite{hamilton2017representation, hamilton2020graph}.  
Define the ground truth, decoder and loss functions by 
\begin{alignat}{3}
 &\!\gt &&: \mathcal N \times \mathcal N \to \R_{\geq 0},&&\quad\! \gt(i, j) = W_{i j}\,, \\ 
 &\!\dec &&: \R^\ell \times \R^\ell \to \R^+, &&\quad \!\dec(u,v) = \Vert u - v\Vert^2\,, \\ 
 & L && : \R_{\geq 0} \times \R^+ \to \R_{\geq 0},&&\quad L(y, \hat y) = y \hat y\,.
\end{alignat} 
Then the goal is to to find the non-constant encoder 
\beq
\enc(\graph) = \left(z_i^T\right)_{i\in\mathcal N}
\eeq
that minimizes the model's loss $\mathcal L \in \R^+$ up to a scaling factor, where that loss is 
\beq \label{eq:le_loss}
\begin{aligned} 
\mathcal L & = \sum_{i, j \in \mathcal N} L\big(\gt(i, j), \dec(z_i, z_j) \big)  \\
& = \sum_{i, j \in \mathcal N} W_{i j} \Vert z_i - z_j \Vert^2\,,   
\end{aligned} 
\eeq
where the minimization is subject to a constraint that prevents the solution from collapsing to a lower dimension.   
Notice that $W_{ij} \geq 0$ is larger when $i$ and $j$ are adjacent, so~\eqref{eq:le_loss} means that the model is punished during training for having node embeddings of adjacent nodes be far apart.  
In theory, the loss $\mathcal L$ could be zero, but in practice this only happens in idealized or degenerate situations.  
For example, a constant encoder gives the same output vector for every node, so $z_i=a$ for all $i\in\mathcal N$.  This satisfies $\mathcal L = 0$ but is not useful. 

Belkin et~al.~\cite{belkin2003laplacian} provide an optimal solution based on generalized eigenvectors of the graph Laplacian.  
The graph Laplacian is the matrix $\Delta = D - W$, where $D$ is a diagonal matrix defined by $D_{i i} = \sum_j W_{j i}$~\cite{merris1994laplacian}.  
The generalized eigenvectors $f_k\in\R^{\vert \mathcal N \vert}$, where $1 \leq k \leq \vert \mathcal N\vert$,  are the solutions to the equation 
\beq
\Delta f_k = \lambda D f_k\,, 
\eeq
where they are enumerated so that their corresponding eigenvalues are in ascending order: 
\beq
0 = \lambda_0 \leq \lambda_1 \leq \cdots \leq \lambda_{\vert \mathcal N \vert - 1}\,.
\eeq
The smallest eigenvalue, $\lambda_0$, is always zero.  
The encoder's output for node $i$ uses the $\ell$ eigenvectors corresponding to the
smallest eigenvalues, excluding $\lambda_0$:   
\beq
z_i = [f_1(i), \ldots, f_{\ell}(i)]^T\,.  
\eeq 
These eigenvectors are used because, after excluding the trivial constant eigenvector, they are the graph's lowest-frequency nontrivial modes: they vary as smoothly as possible over edges with large weights, thereby preserving local neighborhood structure.

Inspired by the success of Laplacian eigenmaps, several papers define node similarity using inner products in a class of techniques called matrix factorization methods~\cite{ahmed2013distributed, cao2015grarep, ou2016asymmetric}.  
Matrix factorization approaches typically rely entirely on the graph's edge structure.    
The three cited methods differ only in how they define the ground truth function, $\gt$.  
In \cite{hamilton2017representation}, Hamilton et~al.\ reformulate each approach into an encoder-decoder framework: 
\begin{alignat}{3}
& \!\gt &&: \mathcal N \times \mathcal N \to \R^+,&&\quad\text{Method dependent}\\
& \!\dec &&: \R^\ell \times \R^\ell \to \R,&&\quad\! \dec(u,v) = u^T v, \\ 
& L && : \R^+ \times \R \to \R^+,&&\quad  L(y, \hat y) = \frac{1}{2} (y - \hat y)^2.  
\end{alignat} 
Given an encoder $\enc$ that provides the node embeddings, $z_i$,
\beq
\enc(\graph) = \left(z_i^T\right)_{i\in\mathcal N} \in \R^{\vert\mathcal N\vert\times\ell}\,,
\eeq
the loss $\mathcal L \in \R^+$ is 
\beq
\begin{aligned} 
\mathcal L & = \sum_{i, j \in \mathcal N} L\big( \gt(i, j), \dec(z_i, z_j) \big) \\ 
& = \frac{1}{2} \sum_{i, j \in \mathcal N} \big( z_i^T z_j - \gt(i, j) \big)^2\,.  
\end{aligned} 
\eeq 
If $Z = \left(z_i^T\right)$ is the matrix of features in $\R^{\vert \mathcal N \vert\times\ell}$, then the above loss satisfies 
\beq\label{eq:matrix_factor}
\mathcal L = \frac{1}{2} \Vert Z Z^T - S \Vert^2\,,
\eeq
where $S$ is the matrix with entries $S_{i j} = \gt(i, j)$.  
Minimizing $\mathcal L$ means finding a matrix $Z$ that factors the ground truth of matrix $S$ as shown in Equation~\eqref{eq:matrix_factor}, which is why the methods are called matrix factorization methods.  


In~\cite{ahmed2013distributed}, Ahmed et~al.\ introduce the first matrix factorization method.  They define the ground truth function by $\gt(i, j) = A_{ij}$, where $A$ is the adjacency matrix.
Hence, their goal is to find a solution that minimizes the loss  
\beq
\mathcal L = \frac{1}{2} \sum_{i, j \in \mathcal N} \left( z_i^T z_i - A_{ij}\right)^2\,. 
\eeq

More recently Perozzi et~al.~\cite{perozzi2014deepwalk} introduce a method not obviously related to matrix factorization, where they use random walks on a graph as a tool to learn node embeddings that capture the edge structure of larger node neighborhoods in a computationally efficient manner.  
After a random initialization of node features, a stochastic gradient descent algorithm updates features to optimize the information necessary to estimate the probability that two nodes co-occur on the same random walk of a fixed length.  
Grover et~al.~\cite{grover2016node2vec}
improves on~\cite{perozzi2014deepwalk} by adding parameters that determine the algorithm's tendency to explore new nodes and its tendency to return to the starting node.  
Finally \cite{qiu2018network} shows that these random walk methods are implicitly doing matrix factorization.  

Matrix factorization methods have the benefit of working on graphs without attributes.  
On attributed graphs, however, all of the examples of shallow models share several shortcomings~\cite{hamilton2017representation}:  
\begin{enumerate} 
\item  They make insubstantial use of the node attributes during training, so they do not use all available information.  
Moreover, these models tend to define similarity in terms of proximity, 
and consequently they usually produce poor results when adjacent nodes in a graph tend to be dissimilar~\cite{ribeiro2017struc2vec}.  
\item Trained models cannot be applied to unseen nodes without further training.  
This is impractical for dynamic graphs and for graphs that are so large that they cannot fit in memory.  
It also means that a model trained in a setting with a lot of labeled data is not transferrable to an unseen graph in a related domain with sparsely labeled data. 
\item The information is not efficiently stored in the model.  
Each trained model is the collection of node features for the graph, which means model parameters are not shared across nodes.  
In particular, the number of parameters grows linearly with $\vert\mathcal N\vert$, which can create memory challenges for processing on large graphs.  
\end{enumerate} 
The next section discusses more powerful encoder-decoder approaches called graph neural networks, which resolve these shortcomings.

\section{Graph Neural Networks}\label{sect:GNNs}

Graph neural networks have several desirable properties.  
They jointly use node attributes and edge information for training, 
trained models can be applied to unseen graphs, and they apply naturally to both undirected and directed graphs.  

This section focuses on GNNs that have so-called message-passing layers (described below).  
The vast majority of GNNs in the literature have message-passing layers.  

\subsection{Encoder layers}\label{sect:gnn_encoder} 

A typical GNN encoder has three classes of layers (see Figure~\ref{fig:gnn_encoder}): pre-processing layers, message-passing layers, and post-processing layers.  
The pre- and post-processing layers sometimes form the identity map.
\begin{figure}[H]
\centering
\begin{tikzpicture}
\node[rectangle,thick, draw = black, text = black, minimum size=0.3cm] (x-1) at  (0, 2.2) {Input};
\node[rectangle,thick, draw = black, text = black, minimum size=0.3cm] (x0) at  (0, 1.1) {Pre-processing Layers};
\node[rectangle,thick, draw = black, text = black, minimum size=0.3cm] (x1) at  (0, 0) {Message-Passing Layers};
\node[rectangle,thick, draw = black, text = black, minimum size=0.3cm] (x2) at  (0, -1.1) {Post-processing Layers};
\node[rectangle,thick, draw = black, text = black, minimum size=0.3cm] (x3) at  (0, -2.2) {Output};

\draw[thick, -stealth] (x-1) -- (x0);
\draw[thick, -stealth] (x0) -- (x1);
\draw[thick, -stealth] (x1) -- (x2);
\draw[thick, -stealth] (x2) -- (x3);

\end{tikzpicture}
\caption{Three classes of layers in a GNN encoder.}\label{fig:gnn_encoder}
\end{figure}

The pre-processing layers map each node attribute vector $x_i$ to a node feature vector $h_i^{(0)}$ in a computation that does not involve the edges of the graph:
\beq  \label{preout}
\pre :\R^m \to \R^{m_0},  \quad \pre(x_i) = h_i^{(0)}\,.
\eeq
Each layer in the pre-processing stack is a single-layer, fully-connected feedforward neural network with trainable parameters. 

The node features from~\eqref{preout} feed into the message-passing
layers, which are the most important layers for the GNN's
performance~\cite{wu2019simplifying}.  While the pre-processing layers
operate on each node independently, the message-passing layers operate
on the entire graph, so we combine the individual node features $h_i\psup{0}$ into a
feature matrix for the whole graph, $H\psup{0}\in \R^{\vert\mathcal N\vert\times m_0}$.  Let $K$ be the number of message passing layers.  For $0\leq k \leq K-1$, the message-passing layer $(k+1)$ is a map 
\beq
\messagep \left( (\mathcal N, A, H^{(k)}) \right) = H^{(k+1)}
\eeq
from the graph with node feature matrix $H\psup{k}$ to the node feature matrix $H\psup{k+1}\in\R^{\vert\mathcal N\vert\times m_{k+1}}$. 
Message-passing layers produce their output node features by aggregating information from each node's immediate neighbors in the graph.  
Then node features from each successive message-passing layer contain information that has been aggregated over a wider set of nodes than the previous layers.  
The final output
from message passing layers
that aggregate node features over 1-hop neighborhoods is a set of low dimensional node features that summarize information in each node's $K$-hop neighborhood.  
In this way, message-passing layers resemble the highly successful convolutional neural networks for image classification.  

Node features from the message-passing layers subsequently feed into the final, post-processing layers of the GNN, which are, like the pre-processing layers, fully connected feedforward neural networks.
Collectively the post-processing layers define the map
\beq
\post:\R^{m_K} \to \R^\ell,\quad \post(h_i^{(K)}) =  z_i.
\eeq
The combined action of all three groups of layers defines the encoder of the GNN, which produces a node embedding $z_i$ for each node:
\beq
\enc (\graph) = \left(z_i^T\right)_{i\in\mathcal N} \in \R^{\vert\mathcal N\vert \times \ell}.
\eeq

Owing to the importance of the message passing layers, we describe them in greater detail next. Each message-passing layer computes its output using the same process. Let $h_i\psup{k} \in \R^{m_k}$ denote the output from layer $k$ for node $i$.
The output from message passing layer $k+1$ is 
\begin{equation}\label{eq:gnn_form}
h_i^{(k + 1)} = \phi\left(\bigwedge\Bigg(h_i^{(k)}, \bigoplus_{j \in \mathcal N_i} \mu_{i j} \Bigg) \right)\,. 
\end{equation}
The function, $\phi$, sometimes called the update function, is differentiable and has trainable parameters. A common choice is a single layer, fully-connected, feedforward neural network.   
The $\bigoplus$ operation is a permutation-invariant aggregation function that combines several vectors into a single vector of the same dimension.  Examples include element-wise sum, mean or maximum.
The set $\mathcal N_i\subset\mathcal N$ is the 1-hop neighborhood of $i$ (excluding $i$),  
and $\mu_{i j}$ is an expression that describes the interaction of node $i$ with node $j$ and depends on $h_i\psup{k}$ and $h_j\psup{k}$ and sometimes on the graph toplogy.
The function $\bigwedge$ determines how each node interacts with the aggregation of its neighbors, for example by concatenation or by element-wise summation. 
Equation~\eqref{eq:gnn_form} is more concisely written without the function $\bigwedge$, but unlike~\cite{bronstein2021geometric}, we include it because our experiments show the choice of $\bigwedge$ is at least as important to the behavior of the GNN as the expresion $\mu_{i j}$.  

The literature describes three common architecture categories for GNNs: Convolutional, Message-Passing (MP), and Attentional.
The type of a GNN is determined by the choice of the expression, $\mu_{ij}$.
We denote the message-passing category by its initials, MP, to help distinguish it from message-passing layers.  
Our description of each GNN category follows \cite{bronstein2021geometric}.  
This discussion is intended to capture key ideas rather than all subtle similarities and differences between individual models. 

An architecture is in the convolutional category when~\eqref{eq:gnn_form} takes the form:
\beq  \label{eq:gcn_form}
h_i^{(k + 1)} = \phi\left(\bigwedge\Bigg(h_i^{(k)}, \bigoplus_{j \in \mathcal N_i } \tau_{i j} \psi\left(h_j^{(k)}\right)\Bigg)\right),
\eeq
where $\psi$ is a differentiable function that can have trainable parameters, usually an affine linear transformation, $\psi(v)=Cv+d$.  
The coefficients $\tau_{i j}\in\R$ are unlearned weights that usually depend only on the local graph topology and encode the connection strength between pairs of nodes~\cite{defferrard2016convolutional, kipf2016semi, hamilton2017inductive, wu2019simplifying, balcilar2020bridging}.  
If the graph exhibits homophily, meaning that nodes with similar features or the same class label tend to be neighbors in the graph~\cite{zhu2020beyond}, 
then in principal, the fixed weights $\tau_{i j}$ make these models a good choice.   
This occurs, for example, in a social network with users connected by friendship~\cite{aiello2012friendship}.   
On the downside, the rigidness of fixed weights may inhibit their ability to represent the complex relationships that arise in low homophily graphs.  

A GNN in the MP architecture category computes feature vectors by 
\beq  \label{eq:mpnn_form}
h_i^{(k + 1)} = \phi\left(\bigwedge\Bigg(h_i^{(k)},  \bigoplus_{j \in \mathcal N_i } \psi\left(h_i^{(k)}, h_j^{(k)}\right) \Bigg) \right),
\eeq
where $\psi: \R^{2 m_k} \to \R^{m_{k}}$ is a differentiable function with  trainable parameters such as an affine linear transformation.
These are the most expressive of the three flavors of GNNs, which makes them suitable for complex modeling tasks like predicting the properties of molecules or complex dynamical systems~\cite{battaglia2016interaction, pmlr-v70-gilmer17a}.  
However, this flexibility also makes them more challenging to train well compared to those in the convolutional category.   

Finally, a model in the attentional category balances the expressiveness of an MP architecture with the simplicity of a convolutional one:
\beq  \label{eq:gat_form}
h_i^{(k + 1)} = \phi\left(\bigwedge\Bigg(h_i^{(k)}, \bigoplus_{j \in \mathcal N_i} a\left(h_i^{(k)}, h_j^{(k)}\right)  \psi\left(h_j^{(k)}\right) \Bigg) \right),
\eeq
where $a$ is a scalar-valued function that can have trainable parameters along with $\psi$~\cite{monti2017geometric, zhang2018gaan, velivckovic2017graph, brody2022how, kim2021how}.  
For example, in~\cite{brody2022how} the function $a$ is computed by 
\beq\label{eq:attention_mechanism}
a\left(h_i^{(k)}, h_j^{(k)}\right) = \frac{\exp(\alpha_{i j})}{\sum_{m \in \mathcal N_i} \exp(\alpha_{i m})}\,, \quad \textrm{where } \alpha_{i m} = v^T \!\cdot \, \sigma\biggl(W \cdot \left(h_i^{(k)} \operatorname{\big\|} h_m^{(k)}\right)\biggr)\,.
\eeq
Here, $\sigma$ is a nonlinear function,  $\|$ is concatenation, and for $d$ hidden dimensions, $v\in\R^d$ is a vector and $W\in\R^{d\times 2m_k}$ is a matrix.
Because $a$ is scalar-valued, learning interactions involves fewer parameters than for the MP networks.  
This makes attentional networks easier to train and run than the MP networks at the cost of being less expressive.  
On the other hand, attentional networks tend to outperform convolutional ones on low homophily graphs~\cite{kim2021how, zheng2022graph}.  

While the interaction between each node and the aggregation of its neighbors specified by $\bigwedge$ does not define the GNN category, it can significantly affect  GNN behavior.  When $\bigwedge$ is addition, like it is for GCN or GATv2, the sum mixes the node's features with the aggregation of its neighbors.  This tends to improve performance when neighboring node features are similar, and it tends to degrade performance otherwise.  On the other hand, when $\bigwedge$ is concatenation, such as for GraphSAGE, the node's features and the aggregation of its neighbors propagate separately, without mixing.  In this case, $h_i\psup{k}$ and $\bigoplus_{j \in \mathcal N_i} \mu_{i j}$ in ~\eqref{eq:gnn_form} are separate which tends to help when neighboring node features differ but hurt otherwise.    See~\cite{hamilton2017inductive, velivckovic2017graph, brody2022how}. 


\begin{table}[H]
\tablefontsize
\captionof{table}{Aggregation Function ($\bigwedge$) and GNN Category for Some Common GNNs}\label{tab:aggregation_function_examples}
\vspace{-3mm}
\centerline{
  \begin{tabular}{l|cc}\hline
    & \multicolumn{2}{c}{GNN Category}\\
 \multicolumn{1}{c|}{$\bigwedge$} &  Convolutional & Attentional \\  
 \hline
 Sum & GCN & GAT, GATv2  \\ 
 Concatenation & GraphSAGE &   \\ 
\end{tabular}}
\end{table} 

A version of convolutional GNNs also exist for the spectral domain, where an aggregation function operates on the eigenvectors of the graph Laplacian~\cite{bruna2013spectral}.  
Comparing with the convolutional GNNs described above, 
the spectral version may provide richer features, but it also is more memory intensive and does not readily extend to directed graphs nor allow predictions on unseen nodes~\cite{zhang2019graph, ma2019spectral}.  

\subsection{Decoder and loss functions}\label{sect:gt_dec_loss} 

Below are minimalist examples of these components for several graph tasks from Section~\ref{sect:graph_tasks}.  
More sophisticated examples appear in the literature.
As usual, $A$ is the graph's adjacency matrix and $\enc(\graph)=\left(z_i^T\right)_{i\in\mathcal N}$ gives the node embeddings.

Given an integer $C > 0$, the softmax function, $\softmax:\R^C \to \R^C$, is defined along each coordinate of a vector $v = [v_1, \ldots, v_C]^T \in \R^C$ by 
\begin{align}\label{eq:softmax}
\softmax(v)_j = \frac{\exp(v_{j})}{\sum_{k = 1}^{C} \exp(v_{k})}\,.   
\end{align}  
Notice that $\sum_{j = 1}^{C} \softmax(v)_j = 1$.  
 
\begin{enumerate}
\item Node classification\label{itm:node_classification}

Let $C$ be the number of class labels, and let $i \in \mathcal N$ be a node, and $y_i = [y_{i 1}, \ldots , y_{i C}]^T \in \{0, 1\}^C$ be a binary vector. 
If node $i$ is in the $c_{\text{th}}$ class, then $y_{i j} = 1$ if $j = c$, and $y_{i j} = 0$ otherwise.  
Then $y_i$ is the class membership vector for $i$, which is ground truth.  
For a matrix $\Theta \in \R^{C \times \ell}$ with trainable parameters, 
the ground truth, decoder and loss functions are 
\begin{alignat}{3}
& \!\gt &&: \mathcal N \to \{0, 1\}^{C},&&\quad\! \gt(i) = y_i, \\ 
& \!\dec&&: \R^{\ell} \to (0, 1)^{C},&&\quad\!\dec (z_i) = \softmax\left(\Theta z_i \right),  \\  
& L && : \{0, 1\}^{C} \times (0, 1)^{C} \to \R^+,&&\quad L(y, \hat y) = - y^T \log(\hat y),
\end{alignat} 
where $\log(\hat y)$ is element-wise logarithm of the prediction vector $\hat y \in (0, 1)^{C}$, so $L$ is the categorical cross-entropy loss function.  
Then the loss for the network is
\beq
\begin{aligned}
\mathcal L & =  \sum_{i} L\left(\gt(i),  \dec(z_i) \right)  \\ 
           & = - \sum_{i} y_i^T \log(\softmax\left(\Theta z_i \right))\,, \label{eq:CE_loss}
\end{aligned}
\eeq
where the sum is usually over a batch of nodes.  
See~\cite{kipf2016semi, schlichtkrull2018modeling, wu2019simplifying, velivckovic2017graph}.  

\item Link prediction \label{itm:link_prediction}

The sigmoid function, $\sigmoid:\R \to \R$, is defined for $t \in \R$ by 
\beq
\sigmoid(t) = \frac{1}{1 + e^{-t}}. 
\eeq  
Define the decoder, ground truth and loss functions by 
\begin{alignat}{3}
 &\!\gt  &&: \mathcal N \times \mathcal N \to \{0, 1\},&&\quad\! \gt(i, j) = A_{i j}\,, \label{eq:link_pred_1} \\ 
 &\!\dec &&: \R^{\ell} \times \R^{\ell} \to (0, 1),        &&\quad\! \dec(z_i, z_j) = \sigmoid(z_i^T z_j) \,, \label{eq:link_pred_2} \\ 
 & L    &&: \{0, 1\} \times (0, 1) \to \R^+,          && \quad L (y, \hat y) = -y \log(\hat y) - (1 - y) \log(1 - \hat y)\,. \label{eq:link_pred_3}
\end{alignat} 
Then the loss is
\beq
\begin{aligned}
 \mathcal L = \sum_{(i, j)} & L(\gt(i) , \dec(z_i, z_j) ) \\ 
            = -\sum_{(i, j)} & A_{i j}  \log\left(\sigmoid(z_i^T z_j)\right) 
                                    +  (1 - A_{i j}) \log\left(1 - \sigmoid(z_i^T z_j)\right),
\end{aligned} 
\eeq 
where the sum is usually over a batch of node pairs.  
See~\cite{zhang2018link, li2020distance, cai2021line} for more sophisticated examples. 

\item Graph classification 

Graph classification can be done like the node classification example but with one additional step.  
After the encoder produces the node embeddings, apply a global aggregator (e.g., entry-wise addition), which combines all node embeddings produced by the encoder into a single feature vector.  
This feature vector represents the graph and can be converted into a prediction, as done in the node classification example.  

Specifically, consider a set of attributed graphs $\graphset$, and let $\graph\in\graphset$ be a graph with nodes, $\mathcal N_\graph$.
Let the node embeddings of $\graph$ be $z_i^{\graph}\in\R^\ell$ for $i\in\mathcal N_\graph$.  In this example we aggregate them using addition so the aggregated feature vector for $\graph$ is $z_\graph=\sum_{i\in\mathcal N_\graph}z_i^\graph\in\R^\ell$.  
The rest follows like the node classification example.  
Let $C$ be the number of class labels for $\graphset$, and let $y_{\graph} = [y_{\graph 1}, \ldots , y_{\graph C}]^T \in \{0, 1\}^C$ be a binary vector.  
If $\graph$ is in the $c_{\text{th}}$ class, then $y_{\graph j} = 1$ if $j = c$ and is zero otherwise.
Then $y_\graph$ is the class membership vector for $\graph$, which is the ground truth.   
For a matrix $\Theta \in \R^{V\times \ell}$ with trainable parameters, the ground truth, decoder and loss functions are 
\begin{alignat}{3}
& \!\gt &&: \mathcal G \to \{0, 1\}^{C},&&\;\;\! \gt(\graph) =  y_{\graph}, \\ 
& \!\dec&&: \R^\ell \to (0, 1)^{C},&&\;\;\!\dec (z_\graph) = \softmax\left(\Theta z_\graph\right),  \\  
  & L && : \{0, 1\}^{C} \times [0, 1]^{C} \to \R^+,&&\;\; L( y, \hat y) = - y^T \log(\hat y).
\end{alignat} 
Then as in the node classification example, $L$ is the cross-entropy loss function.  
The loss for the network is
\beq
\begin{aligned}
\mathcal L & = - \sum_{\graph} L\left(\gt(\graph),  \dec(z_\graph) \right)  \\ 
           & = - \sum_{\graph} y_\graph^T \log\left(\softmax\left(\Theta z_\graph\right)\right)\,, \label{eq:CE_loss_graph}
\end{aligned}
\eeq
where the sum is over a batch of graphs selected from $\mathcal G$ for model training. 
Global mean and max aggregators are also possible.  
See~\cite{xu2018how, pham2017graph, tailor2021we, wang2022molecular}.  

\item Community detection 

Modularity is one of the most commonly used graph clustering metrics in the literature~\cite{fortunato2016community, newman2006modularity}.  
Suppose we are trying to partition the graph's nodes into $C > 1$ clusters.  
Define 
\beq
\delta(i, j) = 
\left\{
\begin{aligned}
& 1 \ \ \ \ \  \textrm{ if } i \textrm{ and } j \textrm{ belong to the same cluster}\,,  \\ 
& 0 \ \ \ \ \ \textrm{ otherwise} \,.
\end{aligned}
\right.
\eeq
\newcommand{\edges}{\vert\mathcal E\vert}%
Let $\edges$ denote the number of edges of the graph, and let $d_i$ be the degree of node $i$.  
Then the modularity metric is 
\beq
\mathcal Q = \frac{1}{2 \edges} \sum_{i, j \in \mathcal N} \left( A_{i j}  - \frac{d_i d_j}{2 \edges}\right) \delta(i, j)\,.  
\eeq
Hence, $\mathcal Q$ measures the divergence of the number of edges within clusters from what one would expect by random chance, 
that is, if the graph had the same node degrees but its edges were assigned by a uniform distribution.  
See~\cite{van2009random} for the construction of random graphs with given node degrees, a method called the configuration model. 
Note that $\mathcal Q$ can be positive or negative, and a large positive value indicates an unusual number of intra-cluster edges, thereby indicating meaningful community structure.  

Tsitsulin et~al.~\cite{Tsitsulin2020Graph} use the modularity metric to define the loss function for their GNN.  
First, they re-write $\mathcal Q$ in a convenient form for a gradient descent optimization.  
Let $d = (d_i)_{i \in \mathcal N}$ be the node degree vector, 
and let 
\beq
B = A - \frac{{d} {d}^T}{2 \edges}\,.  
\eeq 
Then 
\beq
\mathcal Q = \frac{1}{2 \edges} Tr(U^T B U)\,, 
\eeq
where $U \in \{0, 1\}^{\vert \mathcal N \vert \times C}$ is the cluster assignment matrix: $U_{ic} = 1$ if node $i$ belongs to cluster $c$, and $U_{i c} = 0$ otherwise.  

Next, relax the entries of $U$ by allowing them to take values in the interval [0, 1].  
This way we can apply continuous optimization methods to $\mathcal Q$, which is differentiable with respect to the entries of $U$.  
Specifically, let $\Theta \in \R^{C\times\ell}$ be a learnable parameter matrix.  
Define the decoder by: 
\beq
\begin{aligned} 
\!\dec: \R^\ell  \to [0, 1]^{C}, \quad\! \dec(z_i) = \softmax\left(\Theta z_i\right)\,, 
\end{aligned} 
\eeq
where row $i$ of $U$ is 
\beq
U_i = \dec(z_i) \in [0, 1]^{C}\,.
\eeq 

Next we define the $\gt$ function.  
This is somewhat of a misnomer for community detection problems because it lacks ground truth. However, the $\gt$ function here serves the same purpose: guiding training.  
Define
\beq
\begin{aligned} 
\!\gt : \mathcal N \times \mathcal N \to \R, \quad\! \gt(i, j) = B_{i j} = A_{i j} - \frac{d_i d_j}{2 \edges}\,.
\end{aligned}
\eeq
Then the decoder outputs probability estimates that a node is in a given cluster, which determine meaningful communities when the loss 
\beq
\mathcal L = -\frac{1}{2 \edges} Tr(U^T B U) 
\eeq
has a large negative value.  
Notice that $\mathcal L$ is differentiable, so the graph neural network can be trained in an end-to-end fashion.  
See~\cite{Tsitsulin2020Graph} for implementation details, which include a regularity term not included here.  

We just discussed an unsupervised approach to community detection with graph neural networks, but there is also semi-supervised community detection.  Here, the modeler incorporates knowledge that some nodes must be in the same class (must-link constraints) and some nodes cannot be in the same class (cannot-link constraints)~\cite{ren2019semi}.  
Supervised community detection also exists, but this is not common and may be regarded as a type of node classification problem~\cite{chen2017supervised}.  

\item Temporal node regression 

Our final example illustrates that problems with a time variable can fit within the same framework.  
A common node regression problem is to predict numerical values of traffic speed and volume at sensors located on a road network.  
These models can be complex, but a relatively simple one appears in~\cite{wang2018efficient}.  
For each time $t \in \N$, their ground truth, decoder and loss functions are 
\begin{alignat}{3}
& \!\gt_t  &&: \mathcal N \to \R,&&            \quad\! \gt_t(i) \in \R, \\ 
& \!\dec_t &&: \R^{\ell}  \to \R,&&\quad\! \dec_t(z_i) = \sigma\left(\Theta_t^T z_i  + b_t\right), \\ 
& L   &&: \R \times \R \to \R^+,                  &&\quad L(y, \hat y) = (y - \hat y)^2 ,
\end{alignat} 
where $\sigma$ is a nonlinear function, $\Theta_t\in\R^\ell$ is a vector, and $b$ is a scalar, both trainable.  
The goal is to predict the next $T_{\textrm{max}} > 0$ times steps into the future, so their loss includes an average over those time steps.  
Specifically, their loss is the mean square error loss given by 
\beq
\begin{aligned} 
\mathcal L & = \frac{1}{T_{\textrm{max}} \vert \mathcal N_s \vert} \sum_{t = 1}^{T_{\textrm{max}}}  \sum_{i \in \mathcal N_s}  L( \gt_t(i),  \dec_t(z_i) ) \\ 
& = \frac{1}{T_{\textrm{max}} \vert \mathcal N_s \vert} \sum_{t = 1}^{T_{\textrm{max}}} \sum_{i \in \mathcal N_s} \Big( \gt_t(i) - \sigma\left(\Theta_t^T z_i + b_t\right)\Big)^2\,, 
\end{aligned} 
\eeq
where $\mathcal N_s$ is the set of nodes with sensors.  A loss function defined by mean absolute error is used by the top performing models~\cite{zheng2020gman, li2017diffusion}.  
\end{enumerate} 

\subsection{Learning paradigms}\label{sect:learning_paradigms}

Inductive and transductive learning are the two common learning paradigms for graph neural networks. 
In inductive learning, \emph{no test data} is available during training, whereas in transductive learning, \emph{all test data except the test labels} are available during training. 
Thus, inductive learning for node classification is the usual supervised-learning setting. 
For example, consider a coauthor network, where each node is an author, an edge indicates that two authors have worked together, node features represent keywords from their papers, and the node label is the author’s most active field of study~\cite{shchur2018pitfalls}. 
In inductive learning, we may have a training graph that covers the years 2000–2004 and a separate test graph that covers 2005–2007. 
The test graph is not available during training. 
After training, the model is evaluated by comparing its predictions on the test graph with the known labels, i.e., the authors’ most active fields of study.
In transductive learning, the graph from 2000–2007 is available during training, but the labels of the test nodes, i.e., their most active fields of study, are withheld from training.

\subsection{Challenges}


\subsubsection{Low Homophily}\label{sect:low_homophily}

Homophily is the tendency for connected nodes to have similar
characteristics, labels or features and is one aspect of graph
complexity.  
There are different ways to measure homophily, such as feature homophily, which concerns similarity across features, and edge homophily, which is the fraction of the edges that connect two nodes of the same class~\cite{zhu2020beyond}.  

For node classification problems, adjacent nodes in high homophily graphs tend to have the same labels.
Message passing works well on these graphs because it
aggregates similar classes, which tend to have similar features.  
This amplifies feature patterns and reduces noise. 
On the other hand, in low homophily graphs, adjacent nodes typically have different labels and node features, so neighborhood aggregation in message passing tends to mix classes together, 
which blurs class boundaries in feature space, 
leading to worse predictions.

For surveys considering GNNs on low homophily graphs, see \cite{khanam2023homophily, gong2026survey}.  
One approach for dealing with low homophily is to not treat all the
neighbors equally, but instead skip or select neighbors that will be most useful, \cite{zhu2021graph, zhu2020beyond, pei2020geom}.  
Node positional information can help attentional networks weight nodes
appropriately while aggregating neighborhoods in a message passing layer, \cite{ma2021graph}.  
Other methods group feature vectors of node neighborhoods by similar characteristics, and then aggregate the groups of similar vectors separately, \cite{jing2024h, pmlr-v198-dai22b}. 


\subsubsection{Label Scarcity for Node Classification}

Label scarcity in graphs is common in real applications due to the
high cost of manual labeling, data privacy constraints, and large data
set size, where graphs may contain billions of nodes.
Training models on label scarce graphs tends to create models
that do not generalize well.  
Additionally, the weaker signal in label scarce graphs tends to make model training less efficient. 

Many techniques to address label scarcity assume that adjacent nodes
on the graph have similar characteristics, sometimes called the
homophily assumption.  
In fact, message passing is a smoothing operation that propagates similar information between neighbors, so under the homophily assumption, message passing networks are a good tool for mitigating label scarcity.  
Sometimes, further smoothing can improve node classification
predictions by combining message passing networks with traditional label propagation, 
which is an iterative procedure for assigning labels to unlabeled nodes according to the prevalence of labels in a given node's neighborhood \cite{zhu2002learning}.  
So-called heterophily-aware GNNs do not rely on the homophily assumption.  
Instead, they emphasize structural relationships of node neighborhoods and so reduce the importance of labels \cite{zhu2021graph, zhu2020beyond}.  
Other techniques bypass the homophily assumption using a two-step process: first train an unsupervised model to produce generally informative features and then train a classifier on those feature vectors \cite{velivckovic2018deep, you2020graph}. 

\subsubsection{Oversmoothing}  

Repeated neighborhood aggregations of node features tends to make feature vectors of nodes similar to those of their neighbors.  
At some point, nodes become harder to distinguish and predictions become worse, which is oversmoothing.  
For this reason, GNNs tend to have few layers unless there is some architectural modification from what we described in Section~\ref{sect:gnn_encoder}.  
One such modification is to add skip connections, which preserve information from previous layers by adding or concatenating the earlier feature vectors with those at later layers \cite{pmlr-v119-chen20v, li2019deepgcns}.  
This tends to stabilize the network by allowing each layer to learn a small refinement of previous layers rather than  overwriting previous representations. 
Alternatively, several strategies use normalization layers within the network to preserve node feature diversity and prevent too much collapse \cite{zhao2020pairnorm, zhou2020towards, zhou2021understanding}.  
Another common approach is Jumping Knowledge, which aggregates node features across layers in a way that allows the model to choose an appropriate neighborhood scale (i.e. receptive field) to combine node feature information \cite{xu2018jumping}.  
Surveys on oversmoothing include \cite{rusch2023survey} and \cite{jin2025oversmoothing}.  

\subsubsection{Oversquashing}

Unlike oversmoothing which is mostly a depth effect, oversquashing is mainly a graph topology effect.  
Intuitively, over-squashing occurs when too many signals need to squeeze through too few edges and a bottleneck occurs, which makes nodes insensitive to other far away nodes, \cite{alon2021on}.  

Several approaches address over-squashing \cite{rusch2023survey, OverSquSurvey2025}.  
A simple approach is to increase the capacity of information flow between nodes by increasing the number of hidden dimensions of the GNN \cite{OverSquMesPass2023}.  
Another simple approach is to use attention networks, like GATv2, that
can prioritize certain channels and thus allocate more capacity to
some channels than others.
More sophisticated methods locate bottlenecks using negative edge curvature, which highlights edges between otherwise more weakly connected or distinct regions, and effective resistance, which is large between node pairs that are connected by only a small number of alternative paths \cite{topping2022understanding, nguyen2023Revisiting, black2023understanding}.  
Still other approaches try to bypass bottlenecks using global attention mechanisms that enable nodes to learn from distant ones directly \cite{wu2022advances}. 
See \cite{OverSquSurvey2025} for a survey of oversquashing.

\subsubsection{Explainability} 

For tasks where mistakes can be costly, user trust in GNN inferences is crucial, which in turn requires explanations that are understandable and aligned with how the model actually reasons. In the last ten years, many approaches have been developed for providing explanations in terms of graph structures or feature statistics. Local methods such as GNNExplainer~\cite{GNNExplainer2019} and CF-GNNExplainer~\cite{CFGNNExplainer2022} focus on individual inferences. GNNExplainer identifies a compact subgraph and feature subset that are most informative for an inferred node or edge label. CF-GNNExplainter takes a counterfactual approach and identifies the minimum update to the edge structure necessary to change an inferred node label. In effect the edge changes provide insight into node links crucial to the inference. Other approaches focus on providing explanations for more global GNN behavior. ProtGNN \cite{ProtGNN2022}, unlike earlier methods, is applied during GNN training and modifies the underlying GNN architecture. Prototype vectors are produced for each label class allowing the inference of labels to be interpreted in terms of Euclidean proximity between the prototypes and the underlying node representations. GCFExplainer \cite{GCFExplainer2025} addresses graph classification and aims to find a small set of graphs that can serve as close counterfactuals to many of the input graphs. In effect this small set of graphs illustrates the graph structures pertinent to a larger class of inferences. For a detailed discussion of approaches that provide explanations in terms of graph structures or feature statistics, see surveys \cite{Survey2024} and \cite{ACRSurvey2025}.     

More recently, approaches based on Large Language Models (LLMs) have been developed for providing textual explanations for GNN inferences. GraphXAIN \cite{GraphXAIN2025} addresses node classification and aims to produce coherent narratives describing why a GNN produces inferred node labels. An example output for a ``high salary" label on a node in a graph of basketball players is shown below (taken from Fig 1. in \cite{GraphXAIN2025}).

\begin{quote}
\small
"The Graph Neural Network classified NBA player represented by the target node with index 57 as having a 'High Salary'. Central to this prediction is the player's position as a Small Forward, which was among the most significant features. Players in this position, known for their versatile roles on the court, often command higher salaries, which could contribute to the 'High Salary' prediction. The number of games played by the player also heavily influenced the model, as appearing in more games often signals competence and reliability, traits associated with higher earnings. Additionally, the player's height at 200.6 cm placed them around the middle of the height distribution among players, potentially aiding in their adaptability across positions, thereby making them a valuable team asset."
 \end{quote}     

A similar problem was addressed in \cite{GraphNarrator2025} though a different approach to using an LLM was developed.  Other approaches have focused on producing textual, counterfactual explanations \cite{CMPP2024} and \cite{NLCE2025}.

\section{Experiments}\label{sect:experiments} 

This section complements the previous theoretical sections with
experimental results on node classification and link prediction
in the transduction setting (see Section~\ref{sect:learning_paradigms} for a definition). 
The goal is to describe the behavior of GNNs under several training and dataset conditions.  
Our experiments focus on GCN, GATv2, and GraphSAGE because they are commonly used as benchmarks and many GNN architectures are built on top of them~\cite{pan2018adversarially, velivckovic2018deep, li2021training, zhang2018link, Tsitsulin2020Graph, xu2023teal}.  
Table~\ref{tab:aggregation_function_examples} on
page~\pageref{tab:aggregation_function_examples} summarizes
characteristics  of these GNNs.  
Our experiments include two other graph models: Multilayer Perceptron (MLP), which only uses node features, and DeepWalk, which only uses edges.  

We use thirteen open-source datasets: seven high homophily datasets and six low homophily ones. 
In Section~\ref{sect:low_homophily} we noted that there are different ways to measure homophily.  
In what follows, we mean \emph{edge homophily}, i.e., the fraction of edges that connect two nodes of the same class.   
The high homophily datasets are citation networks (Cora, PubMed, CiteSeer, DBPL~\cite{yang2016revisiting, bojchevski2017deep}), co-purchase networks (AmazonComputers, AmazonPhoto~\cite{shchur2018pitfalls}), and a coauthor network (CoauthorCS~\cite{shchur2018pitfalls}).  
The low homophily datasets are webpage-webpage networks (WikipediaSquirrel, WikipediaChameleon, WikipediaCrocodile, Cornell, Wisconsin~\cite{rozemberczki2021multi, pei2020geom}) and a co-occurrence network (Actor~\cite{pei2020geom}).  
All datasets are homogeneous graphs, which means they have a single
node type (e.g., ``article'') and a single edge type (e.g., ``is cited
by'').
Ten of the datasets are node classification networks.  Three of the
datasets---Squirrel, Chameleon and Crocodile---are node regression
datasets, so we transform them into node classification networks by
partitioning their values into five ranges,
where each range defines a class. 

Homophily and the signal-to-noise ratio (SNR) of node features are two measures of complexity in an attributed graph.  
We use a Fisher-type version of SNR to measure the separation of node features when they are grouped by node classes.  
Suppose there are $C$ node classes, and for each class $c \in \{1, \ldots, C\}$, and let $\Sigma_w$ be the pooled within-class covariance 
To keep $\Sigma_w$ well-conditioned, we use a shrunk covariance $\Sigma_w^{(\alpha)}$, for small $\alpha$, 
and define SNR by 
\begin{equation}\label{eq:snr}
\mathrm{SNR}
= \frac{1}{C^2} \sum_{1 \leq b, c\leq C} (\mu_{b}-\mu_{c})^\top (\Sigma_w^{(\alpha)})^{-1} (\mu_{b}-\mu_{c}), 
\end{equation}
where $\mu_{c_k}$ is the within-class centroid of the node features of nodes in class $c_k$.  
See Supplementary Material, Section~\ref{sect:fisher_snr_stable} for details.
Table~\ref{tab:dataset_homophily_scores} reports these metrics for each dataset.   
\begin{table}
\tablefontsize
  \captionof{table}{\captionsize Homophily and SNR values of datasets}\label{tab:dataset_homophily_scores}
  \vspace{-3mm}
\centerline{
  \begin{tabular}{rrll|rrll}\hline
   & Num & Homo- & &  & Num & Homo- &  \\
    Dataset & Nodes & phily & SNR  & Dataset & Nodes & phily & SNR \\
     \hline
Photo & 7650 & 0.83 & 42.24 & Crocodile & 11631 & 0.25 & 327.69 \\ 
DBLP & 17716 & 0.83 & 13.94 & Chameleon & 2277 & 0.24 & 1,624.13 \\ 
Cora & 2708 & 0.81 & 71.09 & Squirrel & 5201 & 0.22 & 17.78 \\ 
CoauthorCS & 18333 & 0.81 & 260.26 & Actor & 7600 & 0.22 & 3.24 \\ 
PubMed & 19717 & 0.80 & 12.42 & Wisconsin & 251 & 0.20 & 81,019.76 \\ 
Computers & 13752 & 0.78 & 44.14 & Cornell & 183 & 0.13 & 103,016.58 \\ 
CiteSeer & 3327 & 0.74 & 967.96 & & & & \\ 
      \hline
\end{tabular}}
\end{table} 

Unlike computer vision or language models, GNNs are often trained from random initialization rather than fine-tuned from a pre-trained model, because they are relatively small. 
We take the same approach. Using PyTorch Geometric, we run all experiments 25 times. 
Plotted curves in Section 5 show mean values over these runs and are averaged across collections of high- or low-homophily datasets; 
they are intended to summarize qualitative trends. 
For dataset-level node-classification results, 95\% confidence intervals are reported in the Supplementary Material.

\subsection{Baseline node classification performance}\label{sect:off_shelf}
~ \\
The baseline GCN, GATv2 and GraphSAGE architectures are two layer message-passing networks with 16 hidden dimensions and no pre-processing or post-processing layers. DeepWalk also has 16 hidden dimensions, and MLP is a three-layer fully connected network where the first layer has 128 hidden dimensions and the second one has 64.   
In the literature, GAT is a benchmark more often than GATv2, but we use GATv2 because it consistently outperforms GAT in our experiments~\cite{brody2022how}.   
We trained the models for at most 200 epochs, with a learning rate of 0.1 and a train/val/test split.  
We used two training sizes, 80\% or 1\%, and we divided the labels not used for training evenly among the validation and test sets.  

Table~\ref{tab:computation_time} reports end-to-end CPU wall-clock time Cora dataset~\cite{yang2016revisiting}, a dataset often used for benchmarking, including training and validation/test inference. 
We do not separate these here because the table is intended only as a coarse comparison across model families on a common CPU setting.  
The graph convolutional models have comparable processing times as MLP, while the attentional network, GATv2, is somewhat slower.  
The shallow embedding model, DeepWalk, is by far the slowest.  
Runtimes for the neural models can often be substantially reduced with GPU acceleration.
\begin{table}
\tablefontsize
  \captionof{table}{\captionsize Average CPU wall-clock time in minutes on the Cora dataset (training + validation / test inference)}\label{tab:computation_time}
  \vspace{-3mm}
\centerline{
  \begin{tabular}{rcccccc}\hline
& GCN 			& GraphSAGE 		& GATv2 			& MLP 			& DeepWalk  \\
 \hline
Minutes & 2.26 $\pm$ 0.26 	& 2.23 $\pm$ 0.28 	& 4.03 $\pm$ 0.23 	& 2.58 $\pm$ 0.31 	& 57.96 $\pm$ 4.29 \\ 
\hline
\end{tabular}}
\end{table} 

The theory in Section~\ref{sect:gnn_encoder} indicates that more flexible models should do better on low homophily graphs and the more rigid ones should outperform on high homophily graphs.   
This is illustrated in Table~\ref{tab:deepsnap_scores_basic}, which provides node classification accuracy scores for each model under each training condition.  

The models GCN and GATv2 both use sum-based neighbor aggregation (summing over the neighbors and the node itself) as in Equation~\eqref{eq:gnn_form}.  
The difference is that GCN uses fixed, degree-normalized weights, whereas GATv2 uses feature-adaptive attention weights. 
In our experiments on low homophily graphs, GATv2 outperforms GCN, consistent with attention down-weighting uninformative neighbors.  
GraphSAGE aggregates neighbors but concatenates the neighbor summary with the node's own features before applying the transformation $\phi$ (see Equation~(\ref{eq:gcn_form}) and Table~\ref{tab:aggregation_function_examples}).  
Concatenation creates a separate self channel in $\phi$ that reduces reliance on potentially conflicting neighbor messages on low homophily graphs. 
Accordingly, we observe GraphSAGE outperforming GATv2 in that regime.  
On high homophily graphs, where neighbor information aligns with labels and smoothing is beneficial, GCN (and GATv2) surpass GraphSAGE, with GCN performing best in our runs.  

Finally, the advantage of GNNs over an MLP is most pronounced on high homophily datasets with scarce labels, because they can use edges to share information that MLP ignores.  
Deepwalk, which captures structural proximity only, performs comparably to GNNs on high homophily graphs but degrades on low homophily ones.  

\begin{table}
\tablefontsize
  \captionof{table}{\captionsize Average node classification accuracy of off-the-shelf GNN architectures on low and high homophily dataset collections with 1\% and 80\% of node labels for training.}\label{tab:deepsnap_scores_basic}
  \vspace{2mm}
\centerline{
  \begin{tabular}{rcc|cccc}
      \multicolumn{5}{c}{Average Node Classification Accuracy of Default Designs} \\ 
  \hline
		 & \multicolumn{2}{c|}{80\% Training}   & \multicolumn{2}{c}{1\% Training}\\
Model Name & High Homophily &  Low Homophily & High Homophily & Low Homophily \\
     \hline
GCN 		& 85.31 			& 37.69 			& \textbf{72.57} 		& 30.99 \\ 
GATv2 		& \textbf{86.95} 	& 44.89 			& 69.41 				& 30.93 \\ 
GraphSAGE 	& 83.77 			& \textbf{56.69} 	& 65.22 				& \textbf{34.33} \\ 
MLP 			& 81.67 			& 54.97 			& 46.52 				& 33.89  \\ 
DeepWalk 	& 81.11 			& 33.97 			& 69.67 				& 24.70  \\ 
\hline
\end{tabular}}
\end{table} 
 
 At first sight, it seems odd that MLP, which does not use edge information, tends to do better on high homophily graphs than low homophily ones.  
 However, our low homophily datasets are also harder in several other ways. 
Cornell and Wisconsin are much smaller, so the MLP sees far fewer labeled examples. 
Actor and Squirrel have low SNR, making their node attributes difficult to learn from.  
Finally, Chameleon, Crocodile, and Squirrel have relatively large feature dimensions, which further complicates learning in the low-label regime.
Specifically, define the relative attribute dimension for a dataset compared to the high homophily datasets by 
\begin{equation}\label{eq:rel_feat_dim}
r_{\textrm{feat}}(D) = \frac{\textrm{Attribute Dimension for } D}{\textrm{Median attribute size of high homophily group}}. 
\end{equation}
For the Chameleon, Crocodile, and Squirrel datasets, the values of $r_{\textrm{feat}}(D)$ are 2.2, 9.2, 2.2, respectively. 

\subsection{GNN width for node classification}\label{sect:hidden_dims}

In high homophily settings, increasing the hidden dimension $d$ (i.e. width) often helps up to a point: additional channels can represent more label-aligned patterns until gains by this mechanism saturate at a large $d$.
By contrast, on low homophily graphs increasing $d$ does not address the core issue that neighbor aggregation mixes incompatible signals.  
Here, using a smaller $d$ makes it less likely that the network will overfit to noise in the input, whereas a larger $d$ only helps when combined with a mechanism that temper mixing, like attention.  
Empirically, we observe little change in node-classification accuracy as $d$ increases.

\begin{table}
\tablefontsize
  \captionof{table}{\captionsize The average improvement of node classification accuracy for the tuned designs over the default ones on high and low dataset collections, with 1\% and 80\% of node labels for training.}\label{tab:deepsnap_scores}
  \vspace{2mm}
\centerline{
  \begin{tabular}{rc|cc|c}
      \multicolumn{5}{c}{Average Node Classification Accuracy Improvement over Default Designs} \\ 
  \hline
		 & 80\% Training  & 80\% Training  & 1\% Training  & 1\% Training  \\
Model Name & High Homophily &  Low Homophily & High Homophily & Low Homophily \\
Difficulty	 & Easy 			& Medium 	& Medium		 & Hard \\ 
     \hline
GCN-64 & +2.59 		& +1.47 		& +3.57 			& \textbf{+0.97} \\ 
GCN-128 & \textbf{+2.86} & \textbf{+1.65} & +3.89 			& +0.28 \\ 
GCN-256 & +2.67 		& +1.20 		& \textbf{+4.41} 	& +0.78 \\ 
\hline
GATv2-64 & +0.18 			& +0.26 		& +2.72 		& $-0.44$ \\ 
GATv2-128 & \textbf{+0.28} 	& \textbf{+0.66} & \textbf{+2.73} & \textbf{+0.96} \\ 
GATv2-256 & +0.11 			& +0.33 		& +2.16 		& $-0.10$ \\ 
\hline
GraphSAGE-64 & +4.12 			& \textbf{+1.22} & +6.27 		& \textbf{+1.44} \\ 
GraphSAGE-128 & \textbf{+4.59} 	& +0.58 		& +6.67 		& +1.32 \\ 
GraphSAGE-256 & +4.53 			& +0.62 		& \textbf{+7.53} & +0.13 \\ 
\hline
\end{tabular}}
\end{table} 
We divide the dataset/training configurations into three conditions: the \emph{easy} condition is where the graph is high homophily with 80\% training data, the \emph{hard} condition is where the graph is low homophily with only 1\% training data, and the other two configurations are the \emph{medium} difficulty condition. 
The most benefit from tuning the hidden dimensions occurs in medium difficulty conditions.

\subsection{GNN depth for node classification}\label{sect:depth} 

Unlike in the width experiments, all encoders in this section use batch normalization after each intermediate layer. 
We focus on metrics that quantify oversmoothing and oversquashing; batch normalization improves the numerical stability of these metrics and makes trends more comparable across datasets. 

\begin{figure*}[t!]
\centering
  \begin{subfigure}[t]{0.5\textwidth}
  \centering
  \includegraphics[width=6cm]{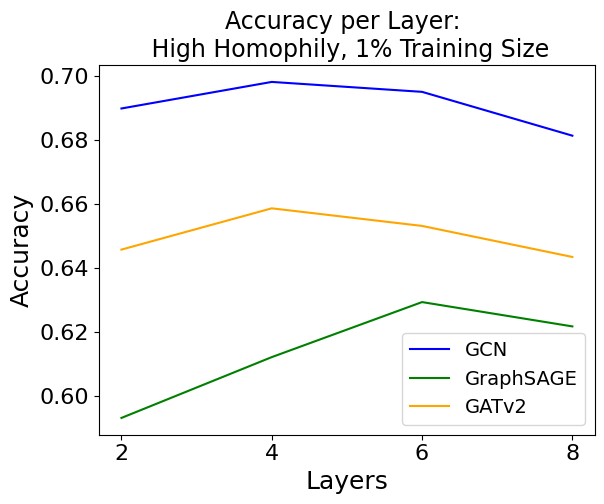}
\caption{}\label{fig:acc_high_homo_a}
\end{subfigure}%
~
  \begin{subfigure}[t]{0.5\textwidth}
  \centering
\includegraphics[width=6cm]{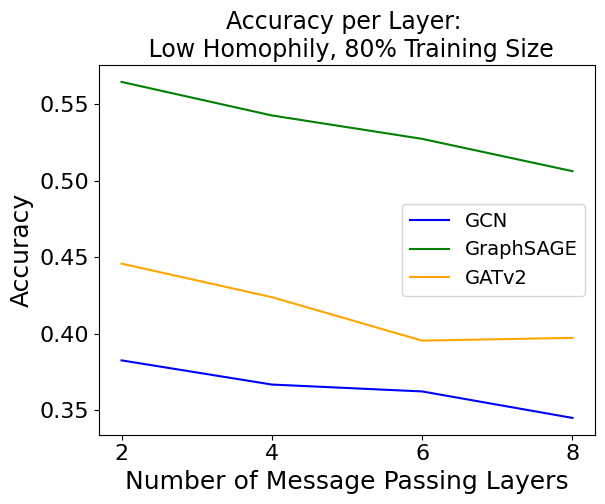}
\caption{}\label{fig:acc_high_homo_b}
\end{subfigure}%
\caption{Test accuracy as a function of GNN depth for three architectures (GCN, GraphSAGE, GATv2). 
Left: high homophily graphs with 1\% of nodes labeled for training. 
Accuracy is relatively stable across depth, with a mild peak around 4–6 layers and decreasing to 8 layers. 
Right: low homophily graphs with 80\% labeled nodes. 
Here accuracy degrades steadily for all models as depth increases.}\label{fig:three_gnns_acc}
\end{figure*}

Figure~\ref{fig:three_gnns_acc} shows test accuracy as a function of the number of message-passing layers: on high homophily graphs, accuracy initially increases with depth and then declines, whereas on low homophily graphs it decreases monotonically. 
This pattern is consistent with prior observations that, without additional architectural modifications, the node-classification accuracy of vanilla GNNs such as GCN, GraphSAGE, and GATv2 degrades once their depth exceeds only a few message-passing layers~\cite{li2018deeper,xu2018jumping}. 
In the next two sections we investigate oversmoothing and oversquashing as complementary explanations for this decline in performance with depth.

\subsubsection{Oversmoothing}\label{sect:num_layers}

The encoder in a node-classification pipeline should produce embeddings whose class means are well separated, while avoiding both excessive within-class variance and low-dimensional collapse (oversmoothing). 
Both phenomena shrink decision margins, but for different reasons. 
In over-dispersion, excessive within-class variance increases overlap between classes, reducing margins. 
In oversmoothing, repeated neighborhood aggregation pushes embeddings into a shared low-dimensional subspace: class means move closer together and within-class variation is confined to only a few directions. 
As a result, many nodes lie near decision boundaries and the remaining directions become overly influential; small noise or bias along these directions is amplified, making nodes harder to distinguish when these directions misalign with class structure.
This intuition is borne out empirically: the 5th-percentile nearest-centroid margins monotonically shrink with depth beyond four layers, 
indicating that an increasing fraction of nodes lie close to decision boundaries (see Supplementary Material, Section~\ref{sect:ncm_q005})

Classical oversmoothing proxies, such as Dirichlet energy of the node features, track the similarity of neighboring node features, but in our experiments (and in line with \cite{Deidda2025RethinkingOversmoothing}) their trends were relatively weak compared to dimension-based diagnostics. 
We therefore quantify collapse using the \emph{Within-Class Effective Rank}, 
which measures how many independent directions of variation remain once class means have been removed; 
lower values indicate stronger oversmoothing.  See Supplementary Material, Section~\ref{sect:within_class_eff_rank} for a formal definition.  

This metric is an instance of the standard entropy-defined effective rank of a matrix \cite{RoyVetterli2007EffectiveRank}, 
which is widely used as an entropy-based measure of effective dimensionality in statistics and deep learning \cite{DelGiudice2021EffectiveDimensionality,Garrido2023RankMe,Zhuo2023RankDifferentialMechanism}. 
Recent work has proposed using the effective or numerical rank of GNN feature representations as an oversmoothing metric \cite{Deidda2025RethinkingOversmoothing}, 
but, to our knowledge, applying effective rank specifically to the covariance of within-class residuals as an oversmoothing diagnostic is novel.

\begin{figure*}[t!]
\centering
  \begin{subfigure}[t]{0.5\textwidth}
  \centering
  \includegraphics[width=6cm]{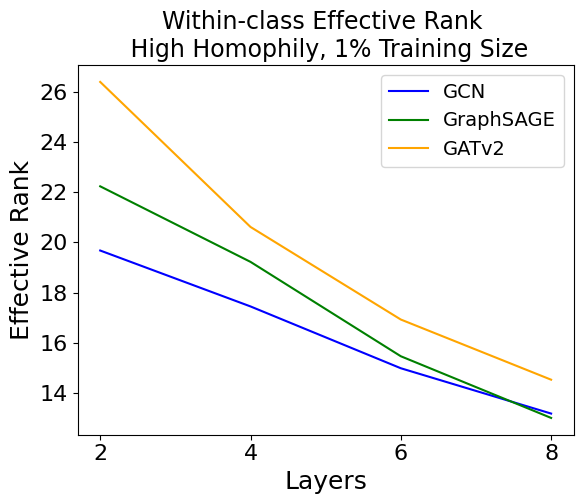}
\caption{}\label{fig:three_gnns_oversmoothing_high_homo_a}
\end{subfigure}%
~
  \begin{subfigure}[t]{0.5\textwidth}
  \centering
\includegraphics[width=6cm]{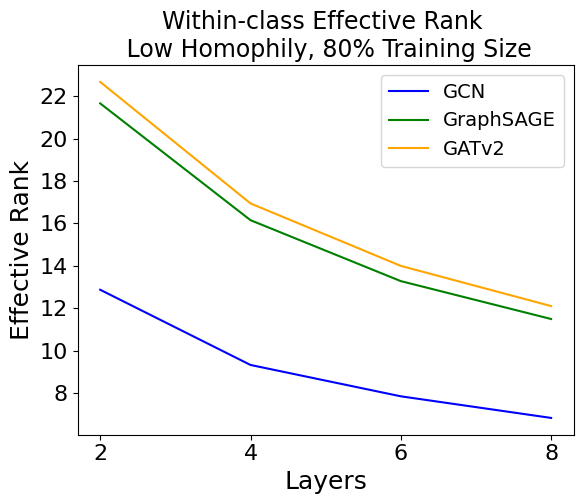}
\caption{}\label{fig:three_gnns_oversmoothing_high_homo_b}
\end{subfigure}%
\caption{Within-class effective rank of node embeddings as a function of GNN depth for three architectures. The within-class effective rank is an entropy-based measure of the effective subspace dimension of the covariance of within-class residuals; lower values indicate stronger representation collapse (oversmoothing). Left: high homophily graphs with 1\% of nodes labeled for training. Right: low homophily graphs with 80\% of nodes labeled for training. In both regimes, effective rank decreases steadily with depth for all models, with GATv2 retaining the most within-class dimensionality, GraphSAGE in between, and GCN collapsing fastest.}\label{fig:three_gnns_oversmoothing_eff_rank}
\end{figure*} 

\paragraph{Comparison of three architectures.}
In Figure~\ref{fig:three_gnns_oversmoothing_eff_rank}, the effective-rank values for all three models, GCN, GATv2, and GraphSAGE, decrease monotonically as the number of layers increases, falling between 6 and 11 units depending on the model.  
This is expected because aggregation in message passing layers are smoothing operations (repeated averaging), which reduces node feature diversity, both globally and within classes. 
The effective rank scores fall accordingly.  

The relative scores align with architectural differences.  
The GCN architecture uses a degree-normalized formula for averaging in its aggregation, which averages rather uniformly.  
Diversity is depressed already at layer two, and it falls at a steady rate.  
The GraphSAGE architecture is different than GCN in that it concatenates a node's own feature vector with an aggregation of its neighbors instead of summing them, see Table~\ref{tab:aggregation_function_examples}, page~\pageref{tab:aggregation_function_examples}.  
This helps it preserve the diversity of the graph's original node feature signal for longer than using pure summation.  
Its effective rank scores therefore trend higher than the rank scores of GCN.  
The GATv2 architecture uses an attention mechanism that can sustain within-class node feature diversity when node features are varied, resulting in high values in shallower networks.  
Its aggregation is also a smoothing operation, causing node feature vectors to look increasingly similar as the layers increase.  
As the inputs to the attention mechanism become more similar, the attention weights also become more similar.  
Aggregation of neighborhood node features then becomes more uniform, behaving closer to that of GCN, leading to a larger drop with depth. 

These mechanisms do not change with the homophily of the graph, so it is expected that we see the same basic shape and model ordering in both the high and low homophily graphs.  
The main difference is scale.  
We observe the largest drop with GCN because it is the least able to avoid mixing incompatible neighborhood node features.  
GraphSAGE and GATv2 scores drop the least because of the self channel and the ability to weight neighbors, respectively, helps them better avoid mixing incompatible feature vectors.  

\begin{figure*}[t!]
\centering
  \begin{subfigure}[t]{0.5\textwidth}
  \centering
  \includegraphics[width=6cm]{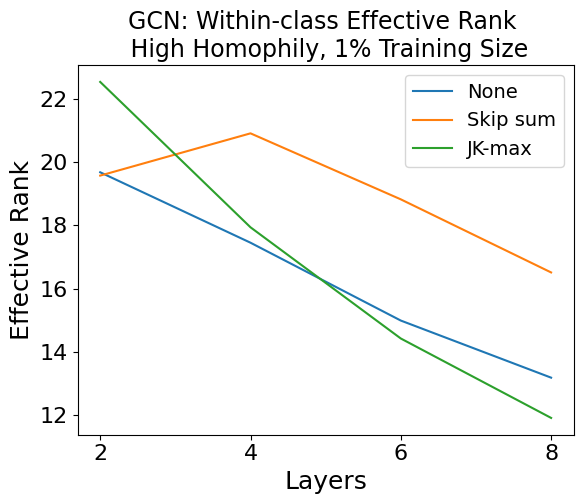}
\caption{}\label{fig:gcn_oversmoothing_high_homo_a}
\end{subfigure}%
~
  \begin{subfigure}[t]{0.5\textwidth}
  \centering
  \includegraphics[width=6cm]{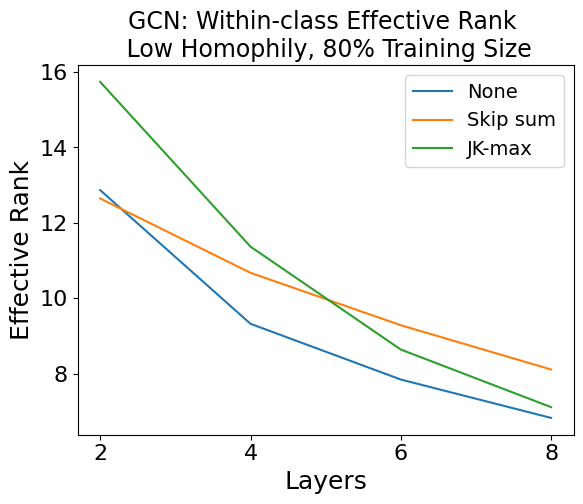}
\caption{}\label{fig:gcn_oversmoothing_low_homo_a}
\end{subfigure}%
\caption{Within-class effective rank of GCN node embeddings as a function of depth for different skip-connection schemes. 
The within-class effective rank is an entropy-based measure of the effective subspace dimension of the covariance of within-class residuals; lower values indicate stronger representation collapse. 
Left: high homophily graphs with 1\% of nodes labeled for training. 
Right: low homophily graphs with 80\% labeled. 
In both regimes, the no-skip model collapses fastest, skip-sum consistently preserves the most within-class dimensionality at larger depths, 
and JK-max starts with the highest rank at shallow depth but quickly converges toward the no-skip baseline as depth increases.}\label{fig:gcn_oversmoothing_eff_rank}
\end{figure*} 

\paragraph{Skip connections.}
Figure~\ref{fig:gcn_oversmoothing_eff_rank} provides plots of GCN under three skip strategies: no skips, residual skip-sum, and Jumping Knowledge (JK).  
Residual skip-sum adds each layer's output to its input, preserving an identify pathway allowing each layer to learn only a refinement of the previous layer's node feature vector instead of an entirely new one \cite{he2016deep, bresson2017residual}.  
We use the maximizing variant of Jumping Knowledge, JK-max, which takes the element-wise maximum feature vector across all layers of the network \cite{xu2018jumping}.  

Residual skip-sum best preserves within-class directions in our experiments.  
Effective rank values grow over the first layers of the network, peaking at four layers, and then decline as repeated aggregations creates redundancy.  
Jumping Knowledge with max pooling preserves sharp differences in node feature vectors because max pooling does not average them away. 
However, as with any architecture that uses message passing, aggregation is a smoothing operation: repeated averaging with neighbors makes node features less sharp and more similar, pulling feature values toward local averages. 
As depth increases, the early layers therefore tend to have sharper features with larger maxima, and Jumping Knowledge keeps selecting those early-layer features, so the final representation relies on a limited set of directions coming predominantly from the early layers. 
Those early-layer representations also continue to smooth during training. 
Empirically, the within-class effective-rank values for Jumping Knowledge eventually fall more quickly than those for the residual skip-sum.
The no-skip method exhibits the strongest smoothing overall.  

\subsubsection{Oversquashing}\label{sect:hidden_oversquashing}

Given a node $n$, we define a Jacobian-based oversquashing diagnostic, Message Squeeze per Bridge Edge (MSBE), as follows.  
Let  
\begin{equation}
B_r(n) = \{u: \mathrm{dist}(u, n) \leq r\}
\end{equation} 
be the radius $r$ ball centered at $n$, where dist is the shortest path distance.  
Let $\mathcal E$ be the set of graph edges and define
\begin{equation}
\partial B_r = \{(u, v) \in \mathcal E: u \in B_r(n)\,, v \notin B_r(n) \ \text{or vice versa}\},
\end{equation} 
be the set of bridge edges crossing the boundary of the ball. 
Then 
\begin{equation}
  \mathrm{MSBE}_r(n)
  =
  \frac{1}{|\partial B_r(n)|+\varepsilon}
  \sum_{u \notin B_r(n)} J(u \to n), 
\end{equation}
where $J(u \to n)$ is the $L^2$ Jacobian influence from a far-away node $u$ to $n$.  
See Section~\ref{sect:msbe_details} for details. 
Intuitively MSBE quantifies, on average, how much far-node signal must traverse each bridge edge in $\partial B_r(n)$ in order to affect $n$.  
Because different nodes and datasets can have very different numbers of boundary-crossing bridge edges, dividing by $|\partial B_r(n)|$ normalizes for local expansion: 
in regions with many bridge edges, the same far-node signal can enter $B_r(n)$ through more routes, 
while in bottlenecked regions with few bridge edges it is concentrated across a smaller cut.

To our knowledge, prior formalizations of oversquashing either use node-to-node sensitivity aggregated over nodes without explicit normalization by boundary size \cite{topping2022understanding, OverSquMesPass2023, black2023understanding} or are model agnostic, purely structural bottleneck measurements such as Cheeger-type expansion, curvature, or effective resistance \cite{banerjee2022oversquashing, topping2022understanding, black2023understanding}.  
MSBE is, to our knowledge, the first model-aware diagnostic that combines node-to-node influence with an expander-style normalization, yielding locally defined measurements that are comparable across datasets.  

\begin{figure*}[t!]
\centering
  \begin{subfigure}[t]{0.5\textwidth}
  \centering
  \includegraphics[width=6cm]{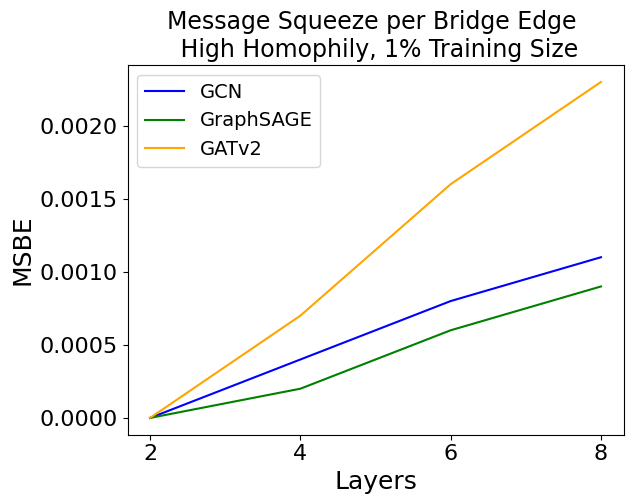}
\caption{}\label{fig:oversquashing_high_homo_a}
\end{subfigure}%
~
  \begin{subfigure}[t]{0.5\textwidth}
  \centering
\includegraphics[width=6cm]{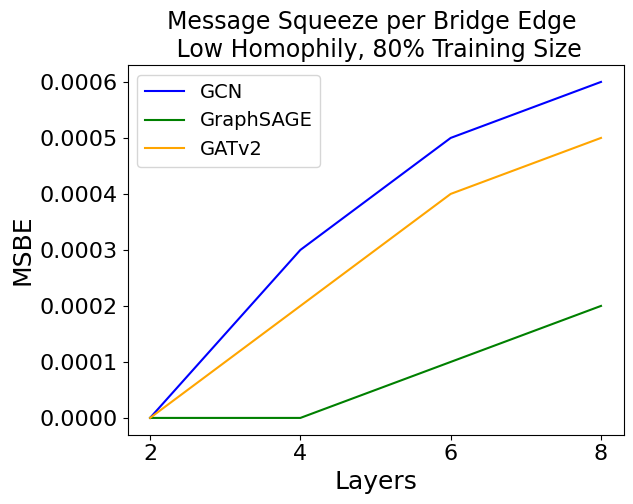}
\caption{}\label{fig:oversquashing_high_homo_b}
\end{subfigure}%
\caption{Message Squeeze per Bridge Edge (MSBE) as a function of GNN depth for GCN, GraphSAGE, and GATv2, averaged over the high homophily graphs with 1\% training labels (left) and the low homophily graphs with 80\% training labels (right). In both regimes, MSBE increases with depth (indicating stronger oversquashing), with GATv2 (respectively GCN) attaining the highest MSBE on high- (respectively low-) homophily graphs and GraphSAGE consistently lowest. MSBE values are noticeably larger in the high homophily setting.}\label{fig:oversquashing}
\end{figure*} 

\paragraph{Experimental results.}
In our experiments, we set $r = 2$, and average MSBE over 256 sampled nodes.  
Notice that for GNNs that have $L > r$ layers, MSBE usually increases with $L$: each additional layer expands the receptive field of $n$ to include more nodes outside $B_r(n)$, while the number of bridge edges $|\partial B_r(n)|$ for the chosen $r$ is fixed.   

The model ordering in Figure~\ref{fig:oversquashing} aligns with the aggregation mechanisms. 
The sum-based update of GCN and GATv2 causes more neighbor information to be imported at each layer, greater cross-boundary flow, and a higher MSBE score compared to GraphSAGE, which uses the concatenation-based update
(See $\bigwedge$ in Table~\ref{tab:aggregation_function_examples} on page~\pageref{tab:aggregation_function_examples}.)  

The qualitative depth trend and ordering are the same in both the high  and low homophily regimes; the difference is primarily in scale. 
In our benchmark, high homophily graphs tend to have fewer cross-neighborhood edges than low homophily graphs, so, all else being equal, the per-edge squeeze (MSBE) is higher.  
On low homophily graphs, the learned architectures often down-weight conflicting neighbor features and rely more on self features. 
This is especially true for GraphSAGE, which preserves a separate self channel via concatenation, and for GATv2, whose attention can reduce neighbor weights.  
The GCN architecture is not flexible enough to reweight individual neighbors, but its learned transformation can shift the relative contribution of self versus neighbors on average, which also tends to reduce neighbor influence in low–homophily graphs.

\paragraph{Improved influence after rewiring.}

To assess mitigation, we use the \emph{Change in Message Squeeze per Bridge Edge} ($\Delta$MSBE).
For a target node $n$, we fix its far set $F_r(n) = \mathcal N \setminus B_r(n)$.
We then add a small number of shortcut edges from $n$ to the $M$ far nodes in $F_r(n)$ with the node-to-node influence $J(u \to n)$.
Intuitively, these are far nodes whose signals already reach $n$ through long, potentially distorted paths, so these are the far nodes most likely to be affected by any bottleneck. 
Adding shortcuts to these nodes opens those routes and reveals how much the graph structure was limiting the model’s ability to use their signal.
Keeping the GNN parameters fixed, we recompute MSBE on the rewired graph and define
\begin{equation}
  \Delta \mathrm{MSBE}_r(n)
  \;=\;
  \frac{
    \displaystyle
    \sum_{u \in F_r(n)} J^{+}(u \to n)
    \;-\;
    \sum_{u \in F_r(n)} J(u \to n)
  }{
    \displaystyle
    |\partial B_r(n)| + \varepsilon
  },
\end{equation}
where $J^{+}(u \to n)$ is the influence on the rewired graph.
A positive \linebreak $\Delta \mathrm{MSBE}_r\!(n)$ 
means that the same far region can now send more signal per bridge edge to $n$, indicating that the shortcuts relieved part of the bottleneck. 
See Supplementary Material, Section~\ref{sect:change_in_msbe} for the formal definition.

\begin{figure*}[t!]
\centering
  \begin{subfigure}[t]{0.5\textwidth}
  \centering 
  \includegraphics[width=6cm]{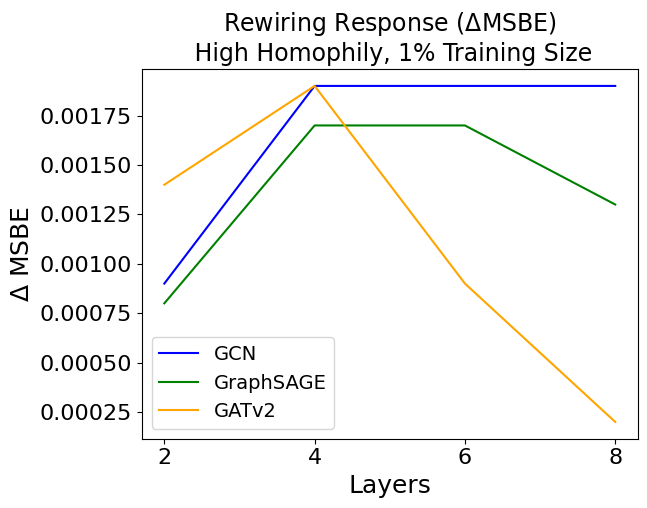}
\caption{}\label{fig:oversquashing_requiring_high_homo}
\end{subfigure}%
~
  \begin{subfigure}[t]{0.5\textwidth}
  \centering
\includegraphics[width=6cm]{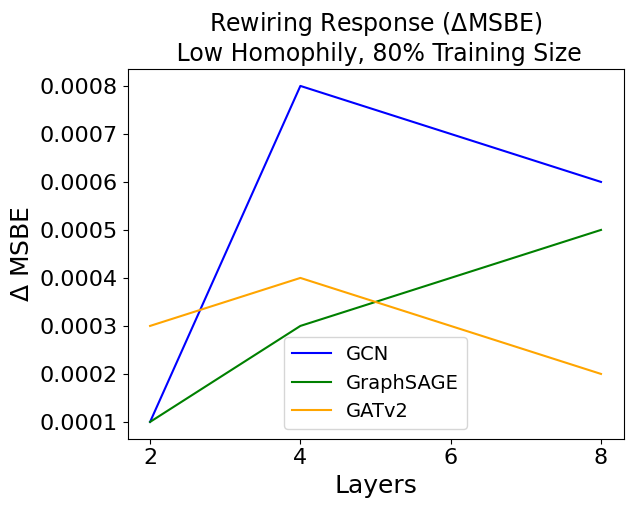}
\caption{}\label{fig:oversquashing_requiring_low_homo}
\end{subfigure}%
\caption{Rewiring response measured by the change in Message Squeeze per Bridge Edge ($\Delta$MSBE) as a function of GNN depth for GCN, GraphSAGE, and GATv2, 
averaged over the high homophily graphs with 1\% training labels (left) and the low homophily graphs with 80\% training labels (right).
For each target node we add four shortcut edges to high-influence far nodes and recompute MSBE; 
positive values indicate that rewiring increases far-node signal per bridge edge.
In both regimes, $\Delta$MSBE usually peaks at moderate depth and then declines, 
showing that the marginal benefit of local rewiring decreases as the
network becomes deeper.}\label{fig:three_gnns_oversmoothing_rewiring}
\end{figure*} 

In our experiments, we compute $\Delta \mathrm{MSBE}_r(n)$ for 128 randomly chosen nodes $n$, using four shortcuts ($M = 4$) and taking far nodes to be at least two hops away ($r = 2$); the results are shown in
Figure~\ref{fig:three_gnns_oversmoothing_rewiring}.
Two opposing effects shape $\Delta \mathrm{MSBE}_r(n)$ as the network depth $L$ increases.
First, deeper networks can reuse the new 1-hop signals from the shortcuts across more layers, 
so $\Delta \mathrm{MSBE}_r(n)$ tends to rise initially.
At the same time, as we add more layers, repeated neighbor-averaging makes node representations smoother: 
nearby nodes look increasingly similar, and changing any single node’s features only slightly perturbs the output at a target node.
In this regime, even a helpful shortcut edge produces a smaller change in far-node influence, so $\Delta \mathrm{MSBE}_r(n)$ eventually peaks and then gradually declines with diminishing returns.

Targeted shortcuts act as persistent 1‑hop inputs reused across layers.  
For GraphSAGE, this occurs without down‑weighting existing high‑weight edges (as attention does in GATv2) and with less global degree-normalization of all neighbors (as in GCN). 
As a result, GraphSAGE tends to show a smaller post-peak decline or a continued but gentler increase with depth---the shortcuts keep paying off, even though the slope ultimately flattens due to the same diminishing-returns effect.  

\subsection{Training sizes for high homophily graphs}\label{sect:training_sizes}

\begin{figure*}[t!]
\centering
  \begin{subfigure}[t]{0.5\textwidth}
  \centering 
  \includegraphics[width=6cm]{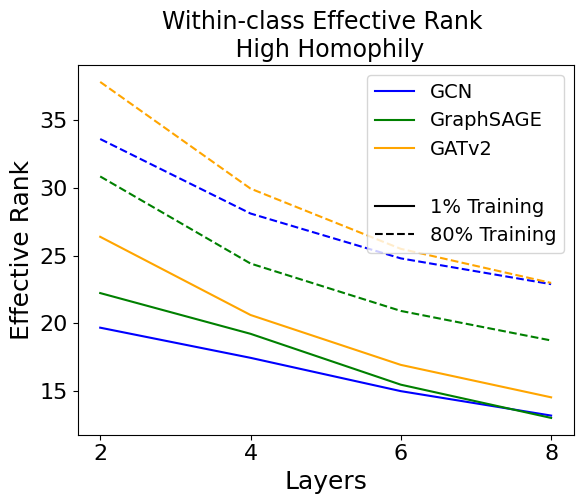}
\caption{}\label{fig:oversmoothing_high_homo_time}
\end{subfigure}%
~
  \begin{subfigure}[t]{0.5\textwidth}
  \centering
\includegraphics[width=6cm]{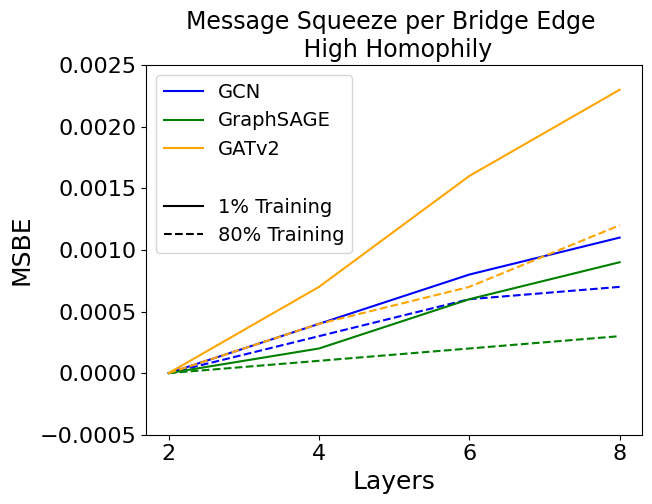}
\caption{}\label{fig:oversquashing_high_homo_time}
\end{subfigure}%
\caption{Within-class effective rank (left) and Message Squeeze per Bridge Edge (MSBE, right) as a function of depth for GCN, GraphSAGE, and GATv2 on high-homophily datasets. Solid lines correspond to models trained with 1\% of node labels and dashed lines to models trained with 80\% of node labels. With more training data, within-class effective rank remains higher, indicating reduced oversmoothing, whereas MSBE tends to be larger in the low-label regime, reflecting stronger oversquashing when supervision is scarce.}\label{fig:three_gnns_time}
\end{figure*} 

To better understand the accuracy gains that come with more training data, we compare our main oversmoothing and oversquashing metrics in the 80\% and 1\% training regimes on high homophily graphs.  
That is, we compare the within-class effective-rank and MSBE metrics.
For low homophily graphs with only 1\% training data, the models perform close to chance, so we omit those results.

The main observation (shown in Figure~\ref{fig:three_gnns_time}) is that the profiles of both metrics as a function of depth are similar for the two training sizes, but their scales differ.
With more training data, the supervised loss provides enough signal to preserve multiple task-relevant directions within each class, increasing the within-class effective-rank values relative to the 1\% training case; this richer class-specific structure is consistent with the higher node-classification accuracy we observe.
For MSBE, with abundant labels—especially on high homophily graphs—the trained models learn to rely less on far-away nodes, which reduces long-range influence and shifts the MSBE profiles downward while preserving their overall shape.
In combination, higher effective rank and lower MSBE indicate embeddings that are better separated and less dominated by noisy long-range messages, explaining the accuracy gains at larger training sizes.

\subsubsection{Improving deeper GNNs with simple architectural tuning}\label{sect:tuning_improvement_1}
In this section, we answer three following questions. 
\\ \\ 
\noindent\textbf{Q1.  Does simple tuning actually improve accuracy?} 
GNNs expose many design variables beyond hidden dimension. 
Rather than using automated hyperparameter search, we focus on a small set of simple, interpretable choices. 
Following You et al.~\cite{you2020design}, we fix the design variables that are shared by top-performing configurations (Table~\ref{tab:model_designs_fixed:1}), and greedily tune the six remaining hyperparameters shown in Table~\ref{tab:model_designs_to_optimize:1}: the numbers of message-passing, post-processing, and pre-processing layers, the skip-connection scheme, the aggregation function, and the learning rate. 
GraphGym~\cite{you2020design}\footnote{\url{https://github.com/snap-stanford/GraphGym.git}} is used to manage experiments.

\begin{table}
\tablefontsize 
  \captionof{table}{\captionsize Hyperparameter values shared by top performing designs in You et~al.~\cite{you2020design}.}\label{tab:model_designs_fixed:1}
  \vspace{-3mm}
\centerline{
  \begin{tabular}{l@{\hspace{10mm}}l@{\hspace{10mm}}l}
    \hline
    Activation (PReLU) & Batch Norm (True) & Dropout (False) \\
    Batch Size (32)    & Optimizer (Adam)  & Epochs (400)    \\
    \hline
  \end{tabular}}
\end{table}

\begin{table}
\tablefontsize 
  \captionof{table}{\captionsize Hyperparameter values to tune}\label{tab:model_designs_to_optimize:1}
  \vspace{-3mm}
\centerline{
  \begin{tabular}{rcll}\hline
  		& 	Tuning	& 	& 	 \\ 
Parameter  &  Order & Starting Value & Options \\
     \hline
 Message-Passing Layers & 1&  2 & 1, 2, 3, 4, 5, 6, 7, 8 \\ 
Post-Processing Layers & 2 & 1 & 1, 2, 3  \\ 
 Pre-Processing Layers & 3 & 1 & 1, 2, 3  \\ 
 Layer Connectivity &  4&  Skip Sum & None, Skip Sum, Skip Concatenate \\
 Aggregation Function  &  5& Mean & Add, Mean, Max  \\
 Learning Rate &  6& 0.01 & 0.005, 0.01, 0.0125, 0.015 \\ 
\end{tabular}}
\end{table} 

For each dataset and architecture, we start from a default configuration with 128 hidden dimensions (Section~\ref{sect:hidden_dims}), train all models for 400 epochs, and select the hyperparameter option in each row of Table~\ref{tab:model_designs_to_optimize:1} that maximizes the average test accuracy over 25 random splits, proceeding in the specified tuning order. 
As in You et al.~\cite{you2020design}, we adjust the number of hidden layers so that different designs have comparable overall size.

\begin{table}
\tablefontsize 
  \captionof{table}{\captionsize The average improvement of node classification accuracy for the tuned designs over the default design with 128 hidden dimensions (see Section~\ref{sect:hidden_dims}) on high and low homophily dataset collections, with 1\% and 80\% of node labels for training. The hyperparameters for each algorithm have been tuned to each dataset.}\label{tab:optimized_versus_default}
  \vspace{2mm}
\centerline{
  \begin{tabular}{rcccc}
      \multicolumn{5}{c}{Average Node Classification Accuracy Improvement over Default with 128 Hidden Dims} \\ 
  \hline
		 & 80\% Training  & 80\% Training  & 1\% Training  & 1\% Training  \\
Model Name & High Homophily &  Low Homophily & High Homophily & Low Homophily \\ 
(Difficulty) 	& (Easy)		& (Medium) 		& (Medium) 	& (Hard) \\
     \hline
GCN 		&  	+0.57		& 	+22.98		&	+0.93	&  $-0.22$\\ 
GATv2 		& 	+1.53		& 	+16.39		& 	+4.57	&  $-0.92$ \\ 
GraphSAGE 	&  	+0.15		&  	+6.18		& 	+4.02	& $-3.89$ \\  
\hline
\end{tabular}}
\end{table}   

\begin{table}
\tablefontsize 
  \captionof{table}{\captionsize The average improvement of node classification accuracy for tuned designs over default ones with 128 hidden dimensions (see Section~\ref{sect:hidden_dims}) on the collection of Cornell and Wisconsin datasets versus the collection of other low homophily datasets, with 80\% of nodes training.  The hyperparameters for each algorithm have been tuned to each dataset.}\label{tab:optimized_versus_default_cor_wis}
  \vspace{2mm}
\centerline{
  \begin{tabular}{rcc} 
\multicolumn{3}{c}{Node Classification Accuracy Improvement} \\ 
  \hline
 80\% Training & Cornell \& Wisconsin &  Other Low Homophily \\
GCN 		& +49.32		& +9.37 	\\ 
GATv2 		& +41.58 		& +3.80	\\ 
GraphSAGE 	& +22.89 		& $-2.12$ 	\\ 
\hline
\end{tabular}}
\end{table} 

Table~\ref{tab:optimized_versus_default} reports the average accuracy improvement of the tuned designs over the default ones across four regimes (high/low edge homophily and 1\%/80\% training labels). 
Tuning provides little benefit in the easiest and hardest settings: when high homophily graphs are already easy (80\% labels), and when low homophily graphs are label-scarce (1\% labels), the tuned and default configurations perform similarly, and in the hardest regime the default can even be slightly better, as for GraphSAGE.  
A likely explanation is that the sweep starts from a generic default configuration, and in the hardest setting the signal is too weak to reliably identify hyperparameter changes that improve on that starting point.
In contrast, tuning yields substantial gains in the medium-difficulty regimes, particularly on low homophily graphs with abundant labels, where GCN and GATv2 improve by roughly 23 and 16 percentage points on average, respectively.  

Much of this gain from low homophily graphs comes from two small, challenging datasets, Cornell and Wisconsin, which high feature SNR. 
Table~\ref{tab:optimized_versus_default_cor_wis} shows that, on these datasets, tuning improves GCN and GATv2 accuracies by 22–49 percentage points, whereas improvements on the other low homophily graphs are more modest. 
These results indicate that simple architectural tuning—adjusting depth, pre-/post-processing layers, and skip connections, along with a few basic training hyperparameters—can significantly improve performance in regimes where the task is neither trivial nor hopeless, while offering limited benefit when the dataset is already easy or severely label-constrained.
\\ \\
\noindent\textbf{Q2. Which architectural components matter?}

\begin{figure*}[t!]
\centering
  \begin{subfigure}[t]{0.5\textwidth}
  \centering
  \includegraphics[width=6cm]{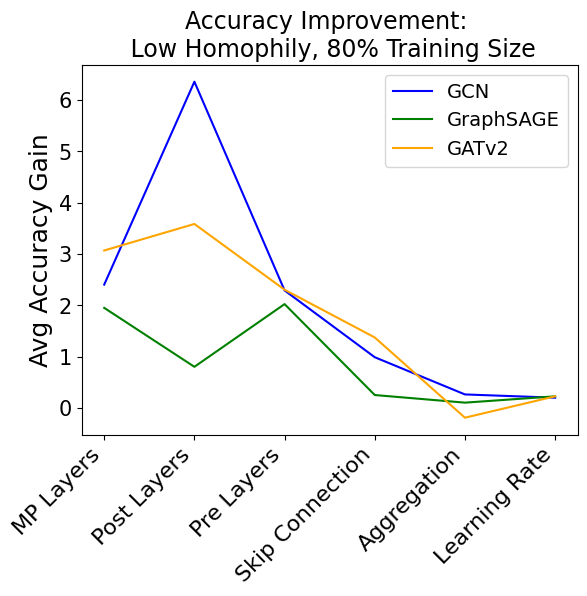}
\caption{}\label{fig:derivative_gain_a}
\end{subfigure}%
~
\begin{subfigure}[t]{0.5\textwidth}
\centering 
\includegraphics[width=6cm]{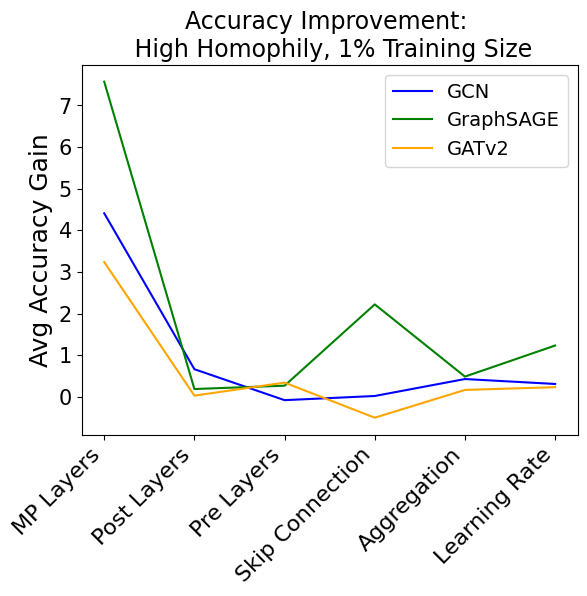}
\caption{}\label{fig:derivative_gain_b}
\end{subfigure}
\caption{Following the greedy hyperparameter tuning process described in Section~\ref{sect:tuning_improvement_1}, the figures show the improved node classification accuracy after tuning the given design component of the GNN compared to the design before it was tuned. }\label{fig:derivative_gain}
\end{figure*}

Figure~\ref{fig:derivative_gain} shows the incremental accuracy gain from tuning each hyperparameter in our greedy procedure. 
Simple structural changes that affect how information is propagated and combined across layers (depth, pre- and post-processing layers, and skip connections) account for most of the achievable gains, whereas tuning the aggregation function and learning rate yields comparatively little improvement. 
Overall, the tuned configurations favor moderate depth, several pre-/post-processing layers, and non-trivial skip connections; we summarize these designs in the Supplementary Material, Section~\ref{sect:gnn_designs}.

\begin{figure*}[t!]
\centering
  \begin{subfigure}[t]{0.5\textwidth}
  \centering 
\begin{tikzpicture}
  \node[anchor=south west, inner sep=0] (img) at (0,0)
      {\includegraphics[width=6cm]{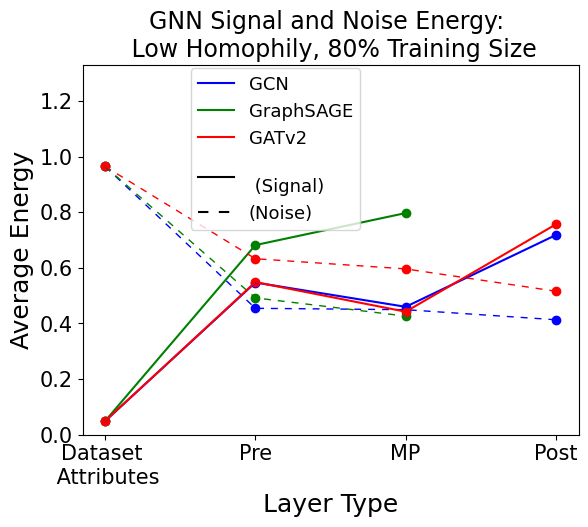}};
  \begin{scope}[x={(img.south east)}, y={(img.north west)}]
      \fill[white] (.15,.875) rectangle (0.95,.98);  
      \node[black] at (.56,.965) {\footnotesize GCN Class Separation};        
      \node[black] at (.56,.905) {\footnotesize High Homophily, 80\% Training Size};        
  \end{scope}
\end{tikzpicture}
\caption{}\label{fig:gnn_snr_a}
\end{subfigure}%
~
  \begin{subfigure}[t]{0.5\textwidth}
  \centering
\begin{tikzpicture}
  \node[anchor=south west, inner sep=0] (img) at (0,0)
      {\includegraphics[width=6cm]{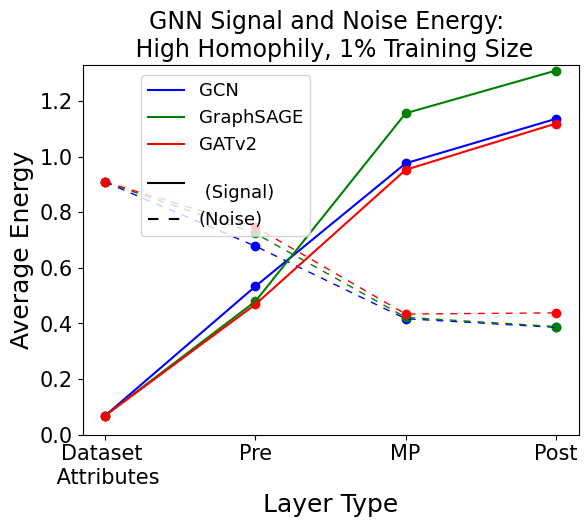}};
  \begin{scope}[x={(img.south east)}, y={(img.north west)}]
      \fill[white] (.15,.875) rectangle (0.95,.98);  
      \node[black] at (.56,.965) {\footnotesize GCN Class Separation};        
      \node[black] at (.56,.905) {\footnotesize Low Homophily, 80\% Training Size};        
  \end{scope}
\end{tikzpicture}
\caption{}\label{fig:gnn_snr_b}
\end{subfigure}%
\caption{GCN between-class distance (“signal”) and within-class variance (“noise”) for each layer type
on the low homophily datasets with 80\% training size.
Higher between-class distance and lower within-class variance indicate better class separation.}\label{fig:signal_noise_variance_per_layer}
\end{figure*}

Figure~\ref{fig:signal_noise_variance_per_layer} illustrates the role of different layer types (pre-/post-processing and message-passing) in terms of between-class distance (“signal”) and within-class variance (“noise”) of node features for the tuned architectures. 
The pre-processing layers transform and rescale the raw node attributes so that nodes from the same class are represented more consistently and within-class variance is reduced before message passing. 
This effect is visible in Figure~\ref{fig:signal_noise_variance_per_layer}: moving from the dataset attributes to the pre-processing layers lowers the noise curves in both the high and low homophily regimes, while keeping the signal at a comparable level. 
The message-passing layers then aggregate these features across neighbors. 
On high homophily graphs, this averaging mainly mixes same-class neighbors, which substantially lowers within-class variance while maintaining or even increasing between-class distance, thereby improving the signal relative to the noise. 
On low homophily graphs, where neighbors are more likely to have different labels, the gain from message passing is more modest, and the tuned architectures therefore favor shallower stacks of message-passing layers. 
Finally, the post-processing layers operate on these refined, propagated features and, in both regimes, produce the highest between-class distance without a corresponding rise in within-class variance, effectively turning the representations into well-separated decision regions.
\\ \\ 
\noindent\textbf{Q3. Comparison with off-the-shelf RevGNN reference models}. 

\begin{table}
\tablefontsize 
  \captionof{table}{\captionsize Node classification accuracy comparison of models on low and high homophily dataset collections, with 1\% and 80\% of node labels for training.}\label{tab:reduced_gnn_with_revgnn}
  \vspace{2mm}
\centerline{
  \begin{tabular}{rcc}
    \multicolumn{3}{c}{Node Classification Accuracy} \\ 
  \hline
 							& 80\% Training  		& 1\% Training  \\ 
Model Name 					&  Low Homophily	 	& High Homophily \\
\hline
GCN$_{\textrm{Tuned}}$			& \textbf{62.32}			& 77.39	\\ 
RevGCN						& 50.93				& \textbf{77.86} \\ 
\hline
GATv2$_{\textrm{Tuned}}$ 		& \textbf{61.94} 		& 76.71	\\ 
RevGATv2					& 53.64				& \textbf{77.99} \\ 
\hline
GraphSAGE$_{\textrm{Tuned}}$ 	& \textbf{63.45} 		& \textbf{75.91} \\ 
RevSAGE 					& 59.61				&  75.87 \\  
\end{tabular}}
\end{table} 

We compare our tuned architectures to off-the-shelf RevGNN variants used in strong Open Graph Benchmark models (Table~\ref{tab:reduced_gnn_with_revgnn}). 
RevGNNs are reversible GNNs built on GCN-, GAT-, or GraphSAGE-style encoders and are designed so that memory usage is essentially independent of depth~\cite{li2021training}, making them an appropriate reference for efficient deep GNN design. 
In our experiments, we use the RevGNN models without dataset-specific tuning, with 160 hidden dimensions, 4 layers, 200 training epochs, and an 80/10/10 train/validation/test split; 
accordingly, this comparison should be interpreted only as a reference against standard RevGNN configurations, not as a controlled comparison between equally tuned model families.  
Under that interpretation, Table~\ref{tab:reduced_gnn_with_revgnn} shows that simple tuning of vanilla GNNs can make them competitive with standard off-the-shelf RevGNN settings.

\subsection{Link Prediction} 

We use an 80/10/10 train/validation/test split with a 50/50 balance between positive and negative edges, and we train for 200 epochs. 
As in the node-classification setting, all results are averages over 25 trials, and we use the same collections of high- and low-homophily datasets.

For link prediction, we remove the validation and test positive edges from the graph and run message passing only on the remaining training graph. 
The GNN encoder then produces a node embedding for each node using the observed graph structure and node attributes. 
To score a candidate edge $(i,j)$, we apply the decoder from Section~4.2, $\sigma(z_i^\top z_j)$, 
where $z_i$ and $z_j$ are the learned node embeddings and $\sigma$ is the sigmoid function. 
Positive examples are observed edges, while negative examples are sampled non-edges. 
During validation and testing, the model is evaluated on held-out positive edges together with sampled negative edges, 
so the target edges are not available to the encoder when constructing the embeddings.

\begin{figure*}[t!]
\centering
  \begin{subfigure}[t]{0.5\textwidth}
  \centering
  \includegraphics[width=6cm]{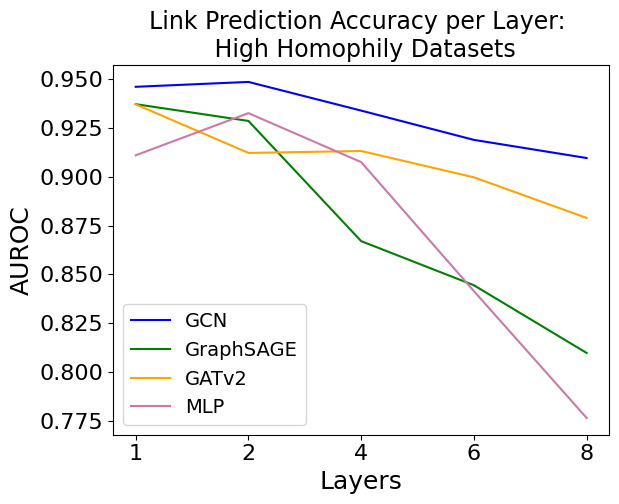}
\caption{}\label{fig:link_pred_depth_high_homo}
\end{subfigure}%
~
  \begin{subfigure}[t]{0.5\textwidth}
  \centering
\includegraphics[width=6cm]{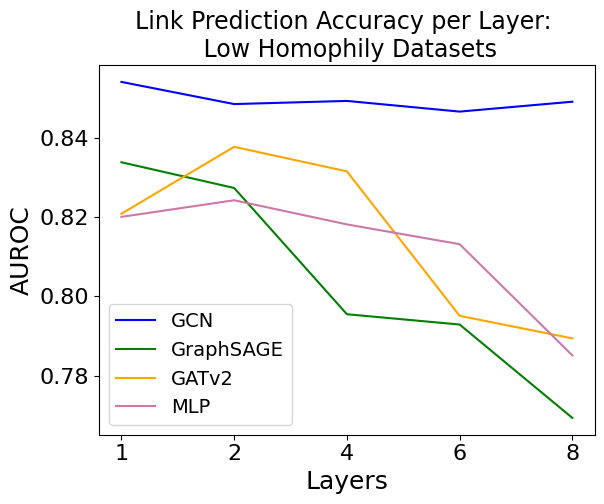}
\caption{}\label{fig:link_pred_depth_low_homo}
\end{subfigure}%
\caption{
Link prediction performance (AUROC) as a function of the number of message-passing layers for GCN, GraphSAGE, GATv2, and an MLP baseline on high and low homophily datasets. 
In both regimes, AUROC typically peaks at 1–2 layers and then degrades with additional depth, consistent with oversmoothing and the propagation of noisy signals in deeper vanilla GNNs, while the MLP, which ignores edges, shows the usual decline in performance for very deep feed-forward networks.}\label{fig:link_pred_depth}
\end{figure*}

\begin{table}
\tablefontsize
\captionof{table}{\captionsize
Effect of pre- and post-processing layers on link prediction AUROC
with 2 message-passing layers. 
$\Delta$AUROC is the improvement over the baseline with no pre- or post-processing layers.}
\label{tab:linkpred_pre_post}
\vspace{-2mm}
\centerline{
\begin{tabular}{l l c c c c}
\hline
Regime      & Model      & Pre Layers & Post Layers & AUROC   & $\Delta$AUROC \\
\hline
High homo.  & GCN        & 0          & 1           & 0.9493  & +0.0095 \\
            & GraphSAGE  & 0          & 1           & 0.9279  & +0.0034 \\
            & GATv2      & 1          & 1           & 0.9408  & +0.0287 \\
\hline
Low homo.   & GCN        & 1          & 1           & 0.8525  & +0.0041 \\
            & GraphSAGE  & 0          & 1           & 0.8318  & +0.0111 \\
            & GATv2      & 0          & 1           & 0.8530  & +0.0253 \\
\hline
\end{tabular}}
\end{table}

With this setup, the depth trends are even more conservative than for node classification (see Figure~\ref{fig:link_pred_depth}). 
Across both high- and low-homophily graphs, performance peaks at around two message-passing layers and then flattens or declines as depth increases.  
Additional layers mix information from increasingly distant parts of the graph; 
this can help node classification by smoothing noisy labels within a class up to moderate depth, 
but for link prediction it tends to wash out the local structural patterns---common neighbors, short paths, and small motifs---that distinguish positive from negative node pairs. 
Once many nodes share very similar embeddings due to oversmoothing, 
the endpoints of an edge and a non-edge can become almost indistinguishable, 
making link prediction especially sensitive to excess depth.

Motivated by this pattern, we next fix the number of message-passing layers at two and tune only the pre- and post-processing layers. 
We find that the best-performing architectures in our experiments always include a single post-processing layer, whereas the benefit of a pre-processing layer is model- and regime-dependent (see Table~\ref{tab:linkpred_pre_post}). 
Two message-passing layers already capture most of the informative local context for link prediction---immediate neighbors and their overlaps, i.e., one- and two-hop structure---so additional preprocessing is mainly helpful when the raw node attributes are noisy or poorly scaled. 
By contrast, the post-processing layer consistently improves performance by adding an extra nonlinearity after local information has been aggregated, 
sharpening the decision boundary in this already well-captured neighborhood regime.

\section{Conclusion} 
A decade ago deep convolutional neural networks for image classification initiated a revolution where feature learning was integrated into the training process of a neural network, and this was subsequently extended to data structures like irregular graphs.  
The encoder-decoder framework neatly describes these models, and the shortcomings of simpler encoder-decoder models motivates the use of more complicated Graph Neural Networks (GNNs).  
Graph neural networks have attracted considerable attention due to state-of-the-art results on a range of graph analysis tasks and datasets, but because of the great variety of graphs and graph analysis tasks, they can be difficult to use for those new to the field.    
As such, we hope our overview of GNNs, their construction and behavior on a variety of datasets and training conditions, has prepared the reader to solve diverse graph problems and understand the technical aspects of literature.  

\section*{Acknowledgements}
This work was partially funded by The MITRE Corporation Independent Research and Development Program.  
After Section 5.2, OpenAI GPT 5-class systems were used to help generate some code and draft some paragraphs, which the authors revised or rewrote. 
OpenAI GPT systems were also used to edit text in Sections 1, 2, 4.4, and 5 and to suggest references for Sections 2 and 5. 
Google Gemini (consumer web/app access) was used to help check the accuracy of some statements.  

~\\
\noindent
\textbf{Approved for Public Release; Distribution Unlimited. Public Release Case Number 26-1038.  ©2026 The MITRE Corporation. ALL RIGHTS RESERVED.}

\bibliographystyle{ACM-Reference-Format}
\bibliography{graph_refs}


\newpage

\appendix

\setcounter{section}{18}
\section{Supplementary Material}

\subsection{Formal definitions of metrics}\label{sect:formulas}

For reference, we have an attributed graph $\graph = \left(\mathcal N, A, \left(x_i^T\right)_{i\in\mathcal N}\right)$.  

\subsubsection{Fisher SNR with eigenvalue-floor shrinkage}\label{sect:fisher_snr_stable}
Suppose that the nodes $\mathcal N$ have $C$ classes.  
Let $\{(x_i,y_i)\}_{i=1}^N$ with $x_i \in \mathbb{R}^d,~ y_i \in \{1,\dots,C\}$.
For class $c$, let $S_c=\{i: y_i=c\},~ n_c=|S_c|$, and
\[
\mu_c = \frac{1}{n_c}\sum_{i\in S_c} x_i, \qquad
\bar\mu = \frac{1}{C}\sum_{c=1}^C \mu_c.
\]
The pooled within-class covariance is
\[
\Sigma_w \;=\; \frac{1}{\sum_{c=1}^C (n_c-1)} \sum_{c=1}^C \sum_{i\in S_c} (x_i-\mu_c)(x_i-\mu_c)^\top.
\]
Let $\Sigma_w = U \Lambda U^\top$ be an eigendecomposition with
\[
\Lambda = \mathrm{diag}(\lambda_1,\dots,\lambda_d), \ \lambda_i\ge 0\,. 
\]
To avoid spurious blow-ups due purely to small-sample noise, we apply a mild eigenvalue floor to the pooled within-class covariance $\Sigma_w$.  
Define $\bar\lambda = \tfrac{1}{d}\mathrm{tr}(\Sigma_w)$.  
For a small $\alpha\in(0,1)$, define the clamped spectrum
\[
\Lambda^{(\alpha)} = \mathrm{diag}\big(\max\{\lambda_1,\alpha\bar\lambda\},\dots,\max\{\lambda_d,\alpha\bar\lambda\}\big),
\]
and the shrunk covariance
\[
\Sigma_w^{(\alpha)} = U \Lambda^{(\alpha)} U^\top.
\]
This guarantees that the shrunk covariance $\Sigma_w^{(\alpha)}$ is well-conditioned.

The stable Fisher signal-to-noise ratio is
\[
\mathrm{SNR}
= \frac{1}{C^2} \sum_{1\le b, c \le C} (\mu_{b}-\mu_{c})^\top (\Sigma_w^{(\alpha)})^{-1} (\mu_{b}-\mu_{c}).
\]

\subsubsection{Within-class Effective Rank}\label{sect:within_class_eff_rank}
Let $\Sigma_w$ be defined as in Section~\ref{sect:fisher_snr_stable}.  
Let $\lambda_1,\dots,\lambda_d \ge 0$ be the eigenvalues of $\Sigma_w$, and define normalized eigenvalue weights
\[
p_i = \frac{\lambda_i}{\sum_{j=1}^d \lambda_j},
\qquad i=1,\dots,d.
\]
Their Shannon entropy (with natural logarithms) is
\[
H(p) = - \sum_{i=1}^d p_i \log p_i.
\]
The \emph{Within-Class Effective Rank} is defined as the entropy-based
effective rank of $\Sigma_w$:
\[
r_{\mathrm{WC}}(\Sigma_w)
= \exp\bigl(H(p)\bigr)
= \exp\!\left(
  - \sum_{i=1}^d p_i \log p_i
\right).
\]
This quantity satisfies $1 \le r_{\mathrm{WC}}(\Sigma_w) \le d$ and can be interpreted as the effective number of independent directions of within-class variation.  

\subsubsection{Nearest-Centroid 5th-Percentile Margin (NCMq0.05)}
\label{sect:ncm_q005}

Let $T, S \subseteq \{1,\dots,N\}$ denote the index sets of training and test nodes, respectively.
For each class $c \in \{1,\dots,C\}$, define the class-specific training set and size
\[
T_c = \{\, i \in T : y_i = c \,\},
\qquad
n_c^{\mathrm{tr}} = |T_c|,
\]
and the corresponding training centroid
\[
\mu_c^{\mathrm{tr}} = \frac{1}{n_c^{\mathrm{tr}}} \sum_{i \in T_c} x_i.
\]

Let $\Sigma_{w,\mathrm{tr}}$ be the pooled within-class covariance computed on the training set $\{x_i\}_{i\in T}$ analogously to $\Sigma_w$ in Section~\ref{sect:fisher_snr_stable}, and let $\Sigma_{w,\mathrm{tr}}^{(\alpha)}$ denote its eigenvalue-floor shrunk version (as in the definition of $\Sigma_w^{(\alpha)}$).
We choose a whitening transform $W \in \mathbb{R}^{d \times d}$ such that
\[
W \,\Sigma_{w,\mathrm{tr}}^{(\alpha)}\, W^\top = I_d.
\]
We then define the whitened embeddings and whitened training centroids by
\[
z_i = W x_i,
\qquad
\tilde\mu_c = W \mu_c^{\mathrm{tr}},
\quad c = 1,\dots,C.
\]

In this whitened space, the nearest-centroid classifier predicts
\[
\hat{y}_i = \arg\min_{c \in \{1,\dots,C\}} \| z_i - \tilde\mu_c \|_2^2.
\]
For a test node $i \in S$ with true class $c = y_i$ and any competing class $j \neq c$, the signed Euclidean distance from $z_i$ to the decision hyperplane between classes $c$ and $j$ (positive on the side of class $c$) is
\[
m_i(c,j)
=
\frac{
(\tilde\mu_c - \tilde\mu_j)^\top z_i
- \tfrac{1}{2}\bigl(\|\tilde\mu_c\|_2^2 - \|\tilde\mu_j\|_2^2\bigr)
}{
\|\tilde\mu_c - \tilde\mu_j\|_2
}.
\]
The nearest-centroid margin of node $i$ is then
\[
m_i = \min_{j \neq c} m_i(c,j), \qquad c = y_i.
\]

\begin{figure*}[t!]
\centering
  \begin{subfigure}[t]{0.5\textwidth}
  \centering 
  \includegraphics[width=6cm]{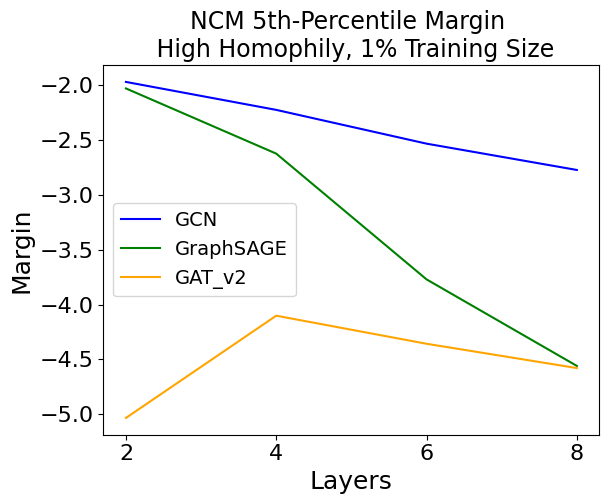}
\caption{}\label{fig:oversmoothing_ncm_high_homo}
\end{subfigure}%
~
  \begin{subfigure}[t]{0.5\textwidth}
  \centering
\includegraphics[width=6cm]{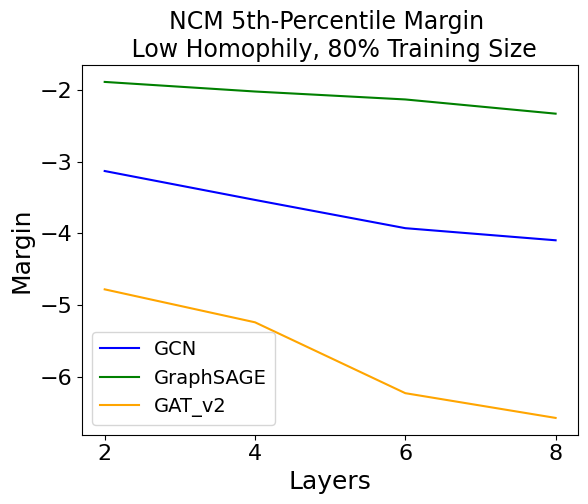}
\caption{}\label{fig:oversmoothing_ncm_low_homo}
\end{subfigure}%
\caption{NCM 5th-percentile margin of node embeddings as a function of GNN depth for three architectures. 
The NCM 5th-percentile margin is the 5th percentile of nearest-centroid margins in a whitened embedding space; lower values indicate that a larger fraction of nodes lie close to decision boundaries. 
Left: high homophily graphs with 1\% of nodes labeled for training. 
Right: low homophily graphs with 80\% of nodes labeled for training. 
In both regimes, margins decrease with depth beyond four layers for all models, indicating that an increasing fraction of nodes lie near decision boundaries as the networks become deeper.}\label{fig:three_gnns_oversmoothing_ncmq5}
\end{figure*}

Let $\{m_i : i \in S\}$ be the collection of margins for all test nodes.
We define the \emph{NCMq0.05} metric as the $5$th-percentile nearest-centroid margin,
\[
\mathrm{NCMq0.05}
=
\operatorname{Quantile}_{0.05}\bigl(\{\, m_i : i \in S \,\}\bigr).
\]
Higher values of NCMq0.05 indicate that the hardest $5\%$ of nodes remain
farther from any decision boundary, whereas lower (more negative) values
indicate that a larger fraction of nodes lie close to or across a decision
boundary.
See Figure~\ref{fig:three_gnns_oversmoothing_ncmq5}, which shows a trend of decreasing margins as the networks get deeper.  

\subsubsection{Message Squeeze per Bridge Edge}\label{sect:msbe_details}

Let $\mathcal E$ be the set of edges in $\graph$.  
Each node $n \in \mathcal N$ has a feature vector $x_n \in \mathbb{R}^d$, and a trained
graph neural network (GNN) produces for each node $n \in \mathcal N$ a scalar output
$\phi(n)$ (for example, the logit of the predicted class at $n$).

\noindent\textbf{1. Distance, ball, boundary, and far set.}
Let $\mathrm{dist}(u,n)$ denote the shortest-path distance between nodes
$u$ and $n$ in $\mathcal N$.
For a fixed node $n \in \mathcal N$ and radius $r < 0$, define the
$r$-hop ball around $n$ by
\begin{equation}
  B_r(n) \;=\; \bigl\{\, u \in \mathcal N : \mathrm{dist}_G(u,n) \le r \,\bigr\}.
\end{equation}
The boundary (set of \emph{bridge edges}) of this ball is
\begin{equation}
  \partial B_r(n)
  \;=\;
  \bigl\{\, (u,v) \in E \;:\;
    u \in B_r(n),\, v \notin B_r(n)
    \ \text{or vice versa}\}.
\end{equation}
The corresponding \emph{far set} is
\begin{equation}
  F_r(n) \;=\; \bigl\{\, u \in \mathcal N : \mathrm{dist}(u,n) > r \,\bigr\}.
\end{equation}

\noindent\textbf{2. Node-to-node influence.}
For a node $u \in \mathcal N$, define the \emph{influence} of $u$ on $n$ via the gradient of $\phi(n)$ with respect to $x_u$.
Let $g(u \to n) \in \mathbb{R}^d$ be this gradient:
\begin{equation}
  g(u \to n)
  \;=\;
  \frac{\partial \phi(n)}{\partial x_u}
  \;\in\; \mathbb{R}^d.
\end{equation}
Let $\sigma \in \mathbb{R}^d$ be a vector of feature-wise standard deviations
(e.g., computed over all node features in the dataset), and let
$\sigma_k > 0$ for $k= 1,\dots,d$. Define the standardized gradient
$\tilde g(u \to n)$ by
\begin{equation}
  \tilde g_k(u \to n)
  \;=\;
  \frac{1}{\sigma_k}\, g_k(u \to n),
  \qquad k = 1,\dots,d.
\end{equation}
Let $d_{\mathrm{eff}} \in \mathbb{N}$ be the number of nonzero entries of
$\tilde g(u \to n)$ (or simply $d$ if all features are used). Then the
\emph{node-to-node influence} is
\begin{equation}
  J(u \to n)
  \;=\;
  \frac{1}{\sqrt{d_{\mathrm{eff}}}}
  \,\bigl\|
    \tilde g(u \to n)
  \bigr\|_2.
\end{equation}

\noindent\textbf{3. Message Squeeze per Bridge Edge.}
For a fixed node $n$ and radius $r$, the \emph{Message Squeeze per Bridge Edge}
(MSBE) at $n$ is defined as
\begin{equation}
  \mathrm{MSBE}_r(n)
  =
  \frac{1}{|\partial B_r(n)|+\varepsilon}
  \sum_{u \in F_r(n)} J(u \to n), 
\end{equation}
where $|\partial B_r(n)|$ is the number of bridge edges of the ball $B_r(n)$
and $\varepsilon > 0$ is a small constant to avoid division by zero.

\subsubsection{Change in Message Squeeze per Bridge Edge}\label{sect:change_in_msbe}
For a fixed node $n \in \mathcal N$ and radius $r > 0$, we consider adding a small set of \emph{shortcut edges} from $n$ into its far set $F_r(n)$.
Given an integer $M \ge 1$, we select a set of $M$ far nodes
$S_r(n) \subseteq F_r(n)$ by taking the $M$ nodes in $F_r(n)$ with the largest influence $J(u \to n)$.  
These define shortcut edges
\begin{equation}
  \mathcal E^{\text{short}}(n)
  \;=\;
  \bigl\{\, (n,u) : u \in S_r(n) \,\bigr\},
\end{equation}
and a corresponding rewired edge set
\begin{equation}
  \mathcal E^{+}(n)
  \;=\;
  \mathcal E \cup \mathcal E^{\text{short}}(n).
\end{equation}
We write $J^{+}(u \to n)$ for the node-to-node influence defined as above, 
but computed on the rewired graph (with edges $\mathcal E^{+}(n))$) with the same trained GNN parameters. 
Then the \emph{change in Message Squeeze per Bridge Edge} at $n$ is then
\begin{equation}
 \Delta\mathrm{MSBE}_r(n)
  \;=\;
  \frac{
    \displaystyle
    \sum_{u \in F_r(n)} J^{+}(u \to n) - \sum_{u \in F_r(n)} J(u \to n)
  }{
    \displaystyle
    |\partial B_r(n)| + \varepsilon
  }.
\end{equation}


\subsection{Architectures Selected by Simple Tuning}\label{sect:gnn_designs} 
The following tables report statistics on the tuned GNN designs from Section~\ref{sect:tuning_improvement_1}.
\begin{table}
\tablefontsize 
\captionof{table}{\captionsize The mean value and $p$-value of the number of
  layers and the learning rates of the hyperparameter tuned models.  
  The hyperparemeters were tuned for each model on the low and high homophily dataset collections with 1\% and 80\% of node labels for training.   
 For each hyperparameter, the $p$-value is for the null hypothesis that the selections over the collection of datasets are drawn from a random distribution.  
 }\label{tab:design_stats:layers_lr}
\vspace{2mm}
\centerline{
  \begin{tabular}{rd{1.2}d{1.2}|d{1.2}d{1.2}|d{1.2}d{1.2}|d{1.4}d{1.2}}
\cc{}&\multicolumn{8}{c}{Statistics of Design Parameters of Tuned GNNs}\\ \cline{2-9}
\cc{80\% Training} & \multicolumn{2}{c|}{Pre Layers} & \multicolumn{2}{c|}{MP  Layers} & \multicolumn{2}{c|}{Post  Layers} & \multicolumn{2}{c}{LR} \\  
\cc{Low Homophily} & \cc{mean}&\ccl{$p$-value}&\cc{mean}&\ccl{$p$-value}&\cc{mean}&\ccl{$p$-value}&\cc{mean}&\cc{$p$-value}\\
\hline
GCN & 2.17& 0.23 & 3.67& 0.21 & 3.0 & 0.0 & 0.024& 0.0 \\ 
GATv2 & 2.0& 0.41 & 3.83& 0.27 & 2.83 & 0.0 & 0.012& 0.17 \\ 
GraphSAGE & 1.83& 0.4 & 6.5& 0.01 & 2.17 & 0.23 & 0.011& 0.34 \\
&&&&&&&&\\
\cc{1\% Training} &&&&&&&&\\
\cc{High Homophily}&&&&&&&&\\
\cline{0-0}
GCN & 1.57& 0.12 & 6.86& 0.0 & 1.43 & 0.05 & 0.010\phantom{0}& 0.37 \\ 
GATv2 & 1.71& 0.25 & 5.86& 0.05 & 1.43 & 0.05 & 0.010\phantom{0}& 0.46 \\ 
GraphSAGE & 2.29& 0.13 & 7.71& 0.0 & 1.71 & 0.25 & 0.0075& 0.02 \\ 
\hline
\end{tabular}}
\end{table}

\begin{table}
\tablefontsize 
\captionof{table}{\captionsize The most common and \% occurring values of the skip connections and aggregation functions on the low and high homophily dataset collections, with 1\% and 80\% of node labels for training.  
The hyperparameters for the algorithms have been tuned to each dataset.  
The \% occurring field says how often the most common selection occurred. }\label{tab:design_stats:skip_agg}
\vspace{2mm}
\centerline{
  \begin{tabular}{rd{0}d{1.1}|d{2.0}d{1.1}}
\cc{}&\multicolumn{3}{c}{Statistics of Design Parameters of Tuned GNNs}\\ \cline{2-4}
\cc{80\% Training} & \multicolumn{2}{c|}{Skip Connections} & \multicolumn{2}{c}{Aggregation} \\  
\cc{Low Homophily} & \cc{most common} & \ccl{\% occurring} &\cc{most common}&\cc{\% occurring}\\
\hline
GCN & \textrm{skip concat} & 66.67 & \textrm{max} & 50.0\phantom{0} \\ 
GATv2 & \textrm{skip concat} & 100.0\phantom{0} & \textrm{add} & 50.0\phantom{0} \\ 
GraphSAGE & \textrm{skip sum} & 66.67 & \textrm{add} & 66.67 \\ 
&&&&\\
\cc{1\% Training} &&&&\\
\cc{High Homophily}&&&&\\
\cline{0-0}
GCN & \textrm{skip sum} & 71.43 & \textrm{max} & 71.43 \\ 
GATv2 & \textrm{skip sum} & 100.0\phantom{0} & \textrm{max} & 57.14 \\ 
GraphSAGE & \textrm{none} & 71.43 & \textrm{add} & 42.86 \\ 
\hline
\end{tabular}}
\end{table} 

\newpage 
\subsection{Tables of baseline GNN results on individual datasets}\label{sect:detailed_results}

All tables report results with 95\% confidence intervals based on 25 runs.  

\begin{table}
\tablefontsize
\captionof{table}{\captionsize The node classification accuracy of default designs (see Table~\ref{tab:deepsnap_scores_basic}).}
\vspace{2mm}
\centerline{
\begin{tabular}{rccc}
\multicolumn{4}{c}{Node Classification Accuracy: 80\% Training and 16 Hidden Dimensions} \\ 
\hline
Dataset & GCN & GATv2 & GraphSAGE \\ 
\hline
Photo & 89.48 $\pm$ 3.47 & 93.55 $\pm$ 0.37 & 77.95 $\pm$ 8.99 \\ 
DBLP & 85.16 $\pm$ 0.29 & 85.11 $\pm$ 0.32 & 84.84 $\pm$ 0.30 \\ 
Cora & 88.28 $\pm$ 0.74 & 85.71 $\pm$ 0.79 & 88.25 $\pm$ 0.66 \\ 
CoauthorCS & 93.22 $\pm$ 0.21 & 91.24 $\pm$ 0.29 & 94.26 $\pm$ 0.21 \\ 
PubMed & 87.21 $\pm$ 0.24 & 86.99 $\pm$ 0.26 & 87.89 $\pm$ 0.38 \\ 
Computers & 77.34 $\pm$ 5.09 & 89.88 $\pm$ 0.39 & 77.45 $\pm$ 5.16 \\ 
CiteSeer & 76.46 $\pm$ 0.93 & 76.19 $\pm$ 0.88 & 75.77 $\pm$ 1.03 \\ 
Crocodile & 61.66 $\pm$ 0.96 & 68.26 $\pm$ 0.69 & 73.26 $\pm$ 0.44 \\ 
Chameleon & 46.90 $\pm$ 1.17 & 55.34 $\pm$ 1.39 & 64.07 $\pm$ 1.30 \\ 
Squirrel & 29.53 $\pm$ 1.07 & 37.68 $\pm$ 0.73 & 45.24 $\pm$ 0.70 \\ 
Cctor & 27.66 $\pm$ 0.68 & 28.89 $\pm$ 0.65 & 35.08 $\pm$ 0.64 \\ 
Wisconsin & 32.00 $\pm$ 3.30 & 40.00 $\pm$ 2.80 & 66.31 $\pm$ 4.05 \\ 
Cornell & 28.42 $\pm$ 2.80 & 39.16 $\pm$ 3.25 & 56.21 $\pm$ 4.11 \\ 
\hline
\end{tabular}}
\end{table}

\begin{table}
\tablefontsize 
\captionof{table}{\captionsize The node classification accuracy of default designs (see Table~\ref{tab:deepsnap_scores_basic}).}
\vspace{2mm}
\centerline{
\begin{tabular}{rccc}
\multicolumn{4}{c}{Node Classification Accuracy: 1\% Training and 16 Hidden Dimensions} \\ 
\hline
Dataset & GCN & GATv2 & GraphSAGE \\ 
\hline
Photo & 78.35 $\pm$ 7.05 & 78.71 $\pm$ 1.81 & 65.22 $\pm$ 5.18 \\ 
DBLP & 74.80 $\pm$ 0.69 & 74.63 $\pm$ 0.89 & 72.50 $\pm$ 0.96 \\ 
Cora & 60.55 $\pm$ 2.78 & 54.89 $\pm$ 2.78 & 55.93 $\pm$ 1.52 \\ 
CoauthorCS & 89.94 $\pm$ 0.22 & 77.13 $\pm$ 0.79 & 86.10 $\pm$ 0.36 \\ 
PubMed & 81.16 $\pm$ 0.37 & 79.14 $\pm$ 0.46 & 78.86 $\pm$ 0.33 \\ 
Computers & 74.83 $\pm$ 4.76 & 72.64 $\pm$ 1.63 & 54.20 $\pm$ 4.22 \\ 
CiteSeer & 48.38 $\pm$ 1.70 & 48.74 $\pm$ 2.26 & 43.70 $\pm$ 1.88 \\ 
Crocodile & 46.44 $\pm$ 1.58 & 45.38 $\pm$ 1.67 & 52.91 $\pm$ 1.30 \\ 
Chameleon & 30.58 $\pm$ 1.59 & 30.56 $\pm$ 1.35 & 31.47 $\pm$ 1.77 \\ 
Squirrel & 22.16 $\pm$ 0.57 & 22.39 $\pm$ 0.66 & 26.93 $\pm$ 1.08 \\ 
Cctor & 24.44 $\pm$ 0.38 & 23.98 $\pm$ 0.49 & 24.48 $\pm$ 0.62 \\ 
Wisconsin & 34.03 $\pm$ 5.74 & 38.60 $\pm$ 4.18 & 40.22 $\pm$ 4.83 \\ 
Cornell & 28.26 $\pm$ 5.58 & 24.65 $\pm$ 5.56 & 29.96 $\pm$ 5.05 \\ 
\hline
\end{tabular}}
\end{table}

\begin{table}
\tablefontsize 
\captionof{table}{\captionsize The node classification accuracy of default designs (see Table~\ref{tab:deepsnap_scores}).}
\vspace{2mm}
\centerline{
\begin{tabular}{rccc}
\multicolumn{4}{c}{Node Classification Accuracy: 80\% Training and 64 Hidden Dimensions} \\ 
\hline
Dataset & GCN & GATv2 & GraphSAGE \\ 
\hline
Photo & 93.69 $\pm$ 0.26 & 93.54 $\pm$ 0.36 & 94.24 $\pm$ 0.54 \\ 
DBLP & 85.69 $\pm$ 0.31 & 85.03 $\pm$ 0.26 & 84.72 $\pm$ 0.36 \\ 
Cora & 88.69 $\pm$ 0.70 & 86.10 $\pm$ 0.78 & 88.59 $\pm$ 0.68 \\ 
CoauthorCS & 93.48 $\pm$ 0.20 & 92.04 $\pm$ 0.33 & 94.69 $\pm$ 0.16 \\ 
PubMed & 87.64 $\pm$ 0.29 & 87.27 $\pm$ 0.39 & 88.23 $\pm$ 0.36 \\ 
Computers & 88.96 $\pm$ 0.42 & 90.27 $\pm$ 0.31 & 88.83 $\pm$ 0.75 \\ 
CiteSeer & 77.09 $\pm$ 0.82 & 75.68 $\pm$ 0.96 & 75.95 $\pm$ 0.68 \\ 
Crocodile & 63.46 $\pm$ 0.75 & 68.93 $\pm$ 0.50 & 73.05 $\pm$ 0.37 \\ 
Chameleon & 47.48 $\pm$ 1.72 & 56.16 $\pm$ 1.23 & 64.21 $\pm$ 0.97 \\ 
Squirrel & 30.20 $\pm$ 0.72 & 38.50 $\pm$ 0.87 & 44.85 $\pm$ 0.77 \\ 
Cctor & 28.59 $\pm$ 0.57 & 28.37 $\pm$ 0.85 & 34.87 $\pm$ 0.65 \\ 
Wisconsin & 36.62 $\pm$ 2.83 & 38.92 $\pm$ 3.82 & 68.62 $\pm$ 3.29 \\ 
Cornell & 28.63 $\pm$ 4.13 & 40.00 $\pm$ 4.25 & 61.89 $\pm$ 4.01 \\ 
\hline
\end{tabular}}
\end{table}

\begin{table}
\tablefontsize 
\captionof{table}{\captionsize The node classification accuracy of default designs (see Table~\ref{tab:deepsnap_scores}).}
\vspace{2mm}
\centerline{
\begin{tabular}{rccc}
\multicolumn{4}{c}{Node Classification Accuracy: 1\% Training and 64 Hidden Dimensions} \\ 
\hline
Dataset & GCN & GATv2 & GraphSAGE \\ 
\hline
Photo & 89.13 $\pm$ 0.69 & 81.97 $\pm$ 1.02 & 80.10 $\pm$ 1.56 \\ 
DBLP & 76.01 $\pm$ 0.50 & 75.08 $\pm$ 0.94 & 73.99 $\pm$ 0.65 \\ 
Cora & 62.96 $\pm$ 2.78 & 61.57 $\pm$ 2.32 & 58.48 $\pm$ 2.77 \\ 
CoauthorCS & 89.81 $\pm$ 0.26 & 80.42 $\pm$ 0.75 & 86.80 $\pm$ 0.55 \\ 
PubMed & 80.98 $\pm$ 0.40 & 79.66 $\pm$ 0.42 & 79.14 $\pm$ 0.38 \\ 
Computers & 83.09 $\pm$ 0.49 & 76.20 $\pm$ 0.81 & 71.25 $\pm$ 3.31 \\ 
CiteSeer & 51.01 $\pm$ 1.87 & 49.99 $\pm$ 2.21 & 50.67 $\pm$ 2.07 \\ 
Crocodile & 46.40 $\pm$ 1.59 & 47.51 $\pm$ 1.30 & 53.96 $\pm$ 0.65 \\ 
Chameleon & 31.84 $\pm$ 1.19 & 30.03 $\pm$ 1.80 & 33.39 $\pm$ 1.38 \\ 
Squirrel & 22.66 $\pm$ 0.62 & 22.84 $\pm$ 0.57 & 27.80 $\pm$ 0.82 \\ 
Cctor & 23.26 $\pm$ 0.80 & 23.91 $\pm$ 0.66 & 24.78 $\pm$ 0.54 \\ 
Wisconsin & 38.32 $\pm$ 4.71 & 33.56 $\pm$ 4.93 & 41.40 $\pm$ 5.31 \\ 
Cornell & 29.26 $\pm$ 6.03 & 25.09 $\pm$ 5.55 & 33.26 $\pm$ 6.12 \\ 
\hline
\end{tabular}}
\end{table}

\begin{table}
\tablefontsize 
\captionof{table}{\captionsize The node classification accuracy of default designs (see Table~\ref{tab:deepsnap_scores}).}
\vspace{2mm}
\centerline{
\begin{tabular}{rccc}
\multicolumn{4}{c}{Node Classification Accuracy: 80\% Training and 128 Hidden Dimensions} \\ 
\hline
Dataset & GCN & GATv2 & GraphSAGE \\ 
\hline
Photo & 93.87 $\pm$ 0.32 & 93.21 $\pm$ 0.29 & 95.24 $\pm$ 0.37 \\ 
DBLP & 85.69 $\pm$ 0.34 & 85.33 $\pm$ 0.26 & 84.89 $\pm$ 0.22 \\ 
Cora & 88.81 $\pm$ 0.65 & 85.97 $\pm$ 0.72 & 88.29 $\pm$ 0.70 \\ 
CoauthorCS & 93.49 $\pm$ 0.20 & 92.34 $\pm$ 0.29 & 94.73 $\pm$ 0.20 \\ 
PubMed & 88.19 $\pm$ 0.34 & 87.31 $\pm$ 0.29 & 88.65 $\pm$ 0.29 \\ 
Computers & 90.23 $\pm$ 0.31 & 90.27 $\pm$ 0.34 & 89.96 $\pm$ 0.51 \\ 
CiteSeer & 76.91 $\pm$ 0.88 & 76.19 $\pm$ 0.93 & 76.80 $\pm$ 0.94 \\ 
Crocodile & 63.13 $\pm$ 0.53 & 69.28 $\pm$ 0.67 & 73.42 $\pm$ 0.60 \\ 
Chameleon & 47.06 $\pm$ 1.88 & 54.46 $\pm$ 1.33 & 64.14 $\pm$ 1.34 \\ 
Squirrel & 29.16 $\pm$ 0.85 & 39.45 $\pm$ 0.88 & 45.02 $\pm$ 0.73 \\ 
Cctor & 28.42 $\pm$ 0.66 & 29.06 $\pm$ 0.48 & 35.15 $\pm$ 0.54 \\ 
Wisconsin & 36.92 $\pm$ 3.25 & 40.62 $\pm$ 3.28 & 68.00 $\pm$ 2.95 \\ 
Cornell & 31.37 $\pm$ 3.48 & 40.42 $\pm$ 4.20 & 57.89 $\pm$ 3.40 \\ 
\hline
\end{tabular}}
\end{table}

\begin{table}
\tablefontsize 
\captionof{table}{\captionsize The node classification accuracy of default designs (see Table~\ref{tab:deepsnap_scores}).}
\vspace{2mm}
\centerline{
\begin{tabular}{rccc}
\multicolumn{4}{c}{Node Classification Accuracy: 1\% Training and 128 Hidden Dimensions} \\ 
\hline
Dataset & GCN & GATv2 & GraphSAGE \\ 
\hline
Photo & 89.57 $\pm$ 0.63 & 81.27 $\pm$ 0.94 & 80.39 $\pm$ 1.61 \\ 
DBLP & 76.43 $\pm$ 0.69 & 75.21 $\pm$ 0.71 & 74.24 $\pm$ 0.54 \\ 
Cora & 64.08 $\pm$ 2.31 & 60.72 $\pm$ 2.97 & 58.55 $\pm$ 2.20 \\ 
CoauthorCS & 89.76 $\pm$ 0.24 & 81.40 $\pm$ 0.84 & 86.25 $\pm$ 0.50 \\ 
PubMed & 81.06 $\pm$ 0.33 & 79.19 $\pm$ 0.36 & 79.54 $\pm$ 0.28 \\ 
Computers & 83.53 $\pm$ 0.50 & 75.27 $\pm$ 1.04 & 72.91 $\pm$ 2.27 \\ 
CiteSeer & 50.80 $\pm$ 1.88 & 51.95 $\pm$ 1.86 & 51.30 $\pm$ 2.08 \\ 
Crocodile & 46.90 $\pm$ 1.72 & 47.44 $\pm$ 2.25 & 54.55 $\pm$ 0.85 \\ 
Chameleon & 31.24 $\pm$ 1.63 & 30.40 $\pm$ 1.36 & 34.75 $\pm$ 1.59 \\ 
Squirrel & 23.46 $\pm$ 0.51 & 22.36 $\pm$ 0.67 & 28.84 $\pm$ 0.77 \\ 
Cctor & 24.11 $\pm$ 0.64 & 23.71 $\pm$ 0.62 & 25.15 $\pm$ 0.45 \\ 
Wisconsin & 33.30 $\pm$ 5.34 & 36.13 $\pm$ 5.30 & 34.29 $\pm$ 5.57 \\ 
Cornell & 28.61 $\pm$ 5.98 & 31.30 $\pm$ 5.87 & 36.30 $\pm$ 4.84 \\ 
\hline
\end{tabular}}
\end{table}

\begin{table}
\tablefontsize 
\captionof{table}{\captionsize The node classification accuracy of default designs (see Table~\ref{tab:deepsnap_scores}).}
\vspace{2mm}
\centerline{
\begin{tabular}{rccc}
\multicolumn{4}{c}{Node Classification Accuracy: 80\% Training and 256 Hidden Dimensions} \\ 
\hline
Dataset & GCN & GATv2 & GraphSAGE \\ 
\hline
Photo & 93.77 $\pm$ 0.24 & 93.46 $\pm$ 0.37 & 94.94 $\pm$ 0.30 \\ 
DBLP & 85.33 $\pm$ 0.33 & 84.78 $\pm$ 0.38 & 84.78 $\pm$ 0.36 \\ 
Cora & 88.13 $\pm$ 1.00 & 86.69 $\pm$ 0.79 & 88.35 $\pm$ 0.91 \\ 
CoauthorCS & 93.37 $\pm$ 0.28 & 92.34 $\pm$ 0.33 & 94.57 $\pm$ 0.20 \\ 
PubMed & 88.31 $\pm$ 0.30 & 87.10 $\pm$ 0.26 & 88.52 $\pm$ 0.21 \\ 
Computers & 90.33 $\pm$ 0.36 & 90.06 $\pm$ 0.35 & 90.56 $\pm$ 0.32 \\ 
CiteSeer & 76.59 $\pm$ 0.66 & 75.02 $\pm$ 1.00 & 76.40 $\pm$ 0.96 \\ 
Crocodile & 64.04 $\pm$ 0.66 & 69.49 $\pm$ 0.54 & 73.29 $\pm$ 0.51 \\ 
Chameleon & 47.13 $\pm$ 1.35 & 56.14 $\pm$ 1.35 & 63.44 $\pm$ 1.21 \\ 
Squirrel & 29.07 $\pm$ 0.62 & 38.93 $\pm$ 0.85 & 45.93 $\pm$ 0.76 \\ 
Cctor & 28.13 $\pm$ 0.56 & 28.71 $\pm$ 0.70 & 35.62 $\pm$ 0.60 \\ 
Wisconsin & 34.92 $\pm$ 3.49 & 38.46 $\pm$ 3.25 & 68.77 $\pm$ 2.77 \\ 
Cornell & 30.11 $\pm$ 3.33 & 39.58 $\pm$ 5.06 & 56.84 $\pm$ 3.20 \\ 
\hline
\end{tabular}}
\end{table}

\begin{table}
\tablefontsize 
\captionof{table}{\captionsize The node classification accuracy of default designs (see Table~\ref{tab:deepsnap_scores}).}
\vspace{2mm}
\centerline{
\begin{tabular}{rccc}
\multicolumn{4}{c}{Node Classification Accuracy: 1\% Training and 256 Hidden Dimensions} \\ 
\hline
Dataset & GCN & GATv2 & GraphSAGE \\ 
\hline
Photo & 89.08 $\pm$ 0.69 & 78.98 $\pm$ 1.62 & 81.93 $\pm$ 1.45 \\ 
DBLP & 76.65 $\pm$ 0.61 & 75.29 $\pm$ 0.93 & 74.45 $\pm$ 0.52 \\ 
Cora & 66.21 $\pm$ 2.31 & 62.17 $\pm$ 2.59 & 61.51 $\pm$ 1.83 \\ 
CoauthorCS & 89.64 $\pm$ 0.23 & 81.75 $\pm$ 1.28 & 84.78 $\pm$ 0.98 \\ 
PubMed & 81.21 $\pm$ 0.33 & 79.07 $\pm$ 0.44 & 79.39 $\pm$ 0.37 \\ 
Computers & 83.59 $\pm$ 0.56 & 72.81 $\pm$ 0.88 & 74.99 $\pm$ 1.95 \\ 
CiteSeer & 52.50 $\pm$ 2.28 & 50.94 $\pm$ 2.04 & 52.20 $\pm$ 1.44 \\ 
Crocodile & 47.64 $\pm$ 1.37 & 46.73 $\pm$ 1.63 & 54.07 $\pm$ 0.73 \\ 
Chameleon & 30.94 $\pm$ 1.69 & 32.71 $\pm$ 1.38 & 33.91 $\pm$ 1.35 \\ 
Squirrel & 23.23 $\pm$ 0.53 & 22.62 $\pm$ 0.54 & 28.52 $\pm$ 0.83 \\ 
Cctor & 24.67 $\pm$ 0.42 & 24.42 $\pm$ 0.61 & 25.73 $\pm$ 0.55 \\ 
Wisconsin & 36.22 $\pm$ 5.71 & 33.43 $\pm$ 5.22 & 40.16 $\pm$ 5.84 \\ 
Cornell & 27.91 $\pm$ 5.30 & 25.09 $\pm$ 6.26 & 24.39 $\pm$ 5.77 \\ 
\hline
\end{tabular}}
\end{table}

\begin{table}
\tablefontsize 
  \captionof{table}{\captionsize The node classification accuracy of the tuned designs with 80\% of nodes labeled for training (see Table~\ref{tab:optimized_versus_default}).}
  \vspace{2mm}
\centerline{
  \begin{tabular}{rccc}
  \multicolumn{4}{c}{Node Classification Accuracy with 80\% Training} \\ 
  \hline
Dataset 			& GCN$_{tuned}$ 			& GraphSAGE$_{tuned}$ 	& GATv2$_{tuned}$ \\
     \hline
Photo 			& 95.62 $\pm$ 0.18 		& 95.68 $\pm$ 0.16 			& 95.65 $\pm$ 0.18 \\ 
DBLP 			& 84.86 $\pm$ 0.18 		& 84.59 $\pm$ 0.22 			& 84.89 $\pm$ 0.17 \\ 
Cora 				& 88.00 $\pm$ 0.49 		& 87.81 $\pm$ 0.47 			& 88.26 $\pm$ 0.46 \\ 
CoauthorCS 	& 95.36 $\pm$ 0.12 		& 95.17 $\pm$ 0.14 			& 95.17 $\pm$ 0.13 \\ 
PubMed 			& 90.04 $\pm$ 0.22 		& 89.89 $\pm$ 0.16 			& 89.99 $\pm$ 0.21 \\ 
Computers 		& 91.95 $\pm$ 0.16 		& 91.56 $\pm$ 0.18 			& 91.95 $\pm$ 0.15 \\ 
CiteSeer 			& 75.35 $\pm$ 0.62 		& 74.88 $\pm$ 0.49 			& 75.41 $\pm$ 0.48 \\ 
Crocodile 		& 70.31 $\pm$ 0.36 		& 70.31 $\pm$ 0.34 			& 70.74 $\pm$ 0.34 \\ 
Chameleon 		& 61.20 $\pm$ 0.80 		& 61.01 $\pm$ 0.78 			& 60.66 $\pm$ 0.82 \\ 
Squirrel 			& 40.29 $\pm$ 0.62 		& 42.23 $\pm$ 0.59 			& 40.79 $\pm$ 0.68 \\ 
Actor 				& 35.19 $\pm$ 0.47 		& 35.68 $\pm$ 0.36 			& 35.26 $\pm$ 0.46 \\ 
Wisconsin 		& 87.37 $\pm$ 1.66 		& 89.73 $\pm$ 1.80 			& 87.45 $\pm$ 2.36 \\ 
Cornell 			& 79.57 $\pm$ 2.29 		& 81.95 $\pm$ 2.00 			& 76.76 $\pm$ 1.99 \\ 
 \hline
\end{tabular}}
\end{table}

\begin{table}
\tablefontsize 
  \captionof{table}{\captionsize The node classification accuracy of tuned designs with 1\% of nodes labeled for training (see Table~\ref{tab:optimized_versus_default}).}\label{tab:model_designs_fixed_}
  \vspace{2mm}
\centerline{
  \begin{tabular}{rccc}
  \multicolumn{4}{c}{Node Classification Accuracy with 1\% Training} \\ 
  \hline
Dataset 			& GCN$_{tuned}$ 			& GraphSAGE$_{tuned}$		& GATv2$_{tuned}$ \\
     \hline
Photo 			& 88.71 $\pm$ 0.68 		& 85.98 $\pm$ 0.87 				& 87.96 $\pm$ 0.78 \\ 
DBLP 			& 75.51 $\pm$ 0.60 		& 73.75 $\pm$ 0.74 				& 75.64 $\pm$ 0.60 \\ 
Cora 				& 66.62 $\pm$ 1.52 		& 66.65 $\pm$ 2.15 				& 63.53 $\pm$ 2.16 \\ 
CoauthorCS 	& 90.60 $\pm$ 0.43 		& 89.59 $\pm$ 0.60 				& 90.52 $\pm$ 0.22 \\ 
PubMed 			& 79.71 $\pm$ 0.40 		& 78.06 $\pm$ 0.43 				& 79.50 $\pm$ 0.45 \\ 
Computers 		& 82.54 $\pm$ 0.50 		& 82.08 $\pm$ 0.80 				& 83.51 $\pm$ 0.51 \\ 
CiteSeer 			& 58.01 $\pm$ 1.50 		& 55.32 $\pm$ 1.94 				& 56.32 $\pm$ 1.45 \\ 
Crocodile 		& 56.12 $\pm$ 0.78 		& 57.02 $\pm$ 0.76 				& 55.54 $\pm$ 0.60 \\ 
Chameleon 		& 33.46 $\pm$ 1.39 		& 33.42 $\pm$ 1.22 				& 33.52 $\pm$ 1.38 \\ 
Squirrel 			& 25.88 $\pm$ 0.59 		& 27.70 $\pm$ 0.70 				& 25.80 $\pm$ 0.47 \\ 
Actor 				& 25.65 $\pm$ 0.35 		& 25.82 $\pm$ 0.51 				& 25.66 $\pm$ 0.45 \\ 
Wisconsin 		& 27.93 $\pm$ 0.08 		& 27.69 $\pm$ 0.42 				& 27.69 $\pm$ 0.45 \\ 
Cornell 			& 17.25 $\pm$ 1.23 		& 18.90 $\pm$ 1.52 				& 17.52 $\pm$ 1.30 \\ 
 \hline
\end{tabular}}
\end{table}

\begin{table}
\tablefontsize 
  \captionof{table}{\captionsize The node classification accuracy of RevGNNs with 80\% of node labels for training on low homophily graphs (see Table~\ref{tab:reduced_gnn_with_revgnn}).}
  \vspace{2mm}
\centerline{
  \begin{tabular}{rcccccc} 
    \multicolumn{4}{c}{Node Classification Accuracy of RevGNNs} \\ 
  \hline
Dataset 		& RevGCN 		& RevSAGE 		& RevGATv2 		\\
     \hline
Crocodile 		& 69.01 $\pm$ 0.55 	& 73.10 $\pm$ 0.53 	& 71.36 $\pm$ 0.54 \\ 
Chameleon 	& 54.60 $\pm$ 1.18 	& 64.10 $\pm$ 1.36 	& 60.00 $\pm$ 1.49 \\ 
Squirrel 		& 36.25 $\pm$ 0.90 	& 46.34 $\pm$ 1.07 	& 42.63 $\pm$ 0.80 \\ 
Actor 		& 33.29 $\pm$ 0.61 	& 36.91 $\pm$ 0.73 	& 36.22 $\pm$ 0.67 \\ 
Wisconsin 	& 61.38 $\pm$ 3.50 	& 71.08 $\pm$ 2.73 	& 63.85 $\pm$ 3.28 \\ 
Cornell 		& 51.16 $\pm$ 4.06 	& 66.11 $\pm$ 2.80 	& 47.79 $\pm$ 2.91 \\ 
 \hline
\end{tabular}}
\end{table}

\begin{table}
\tablefontsize 
  \captionof{table}{\captionsize The node classification accuracy of RevGNNs with 1\% of node labels for training on high homophily graphs (see Table~\ref{tab:reduced_gnn_with_revgnn}).}
  \vspace{2mm}
\centerline{
  \begin{tabular}{rcccccc}
      \multicolumn{4}{c}{Node Classification Accuracy of RevGNNs} \\ 
  \hline
Dataset 		& RevGCN 			& RevSAGE 			& RevGATv2 &\\
     \hline
Photo 		& 88.89 $\pm$ 0.86 		& 87.12 $\pm$ 0.81 		& 90.21 $\pm$ 0.54 \\ 
DBLP 		& 77.73 $\pm$ 0.69 		& 76.21 $\pm$ 0.72 		& 77.71 $\pm$ 0.68 \\ 
Cora 		& 67.37 $\pm$ 1.63 		& 62.41 $\pm$ 2.03 		& 67.96 $\pm$ 1.94 \\ 
CoauthorCS 	& 91.59 $\pm$ 0.34 		& 90.72 $\pm$ 0.29 		& 90.86 $\pm$ 0.34 \\ 
PubMed 		& 81.83 $\pm$ 0.39 		& 79.98 $\pm$ 0.36 		& 81.02 $\pm$ 0.45 \\ 
Computers 	& 83.19 $\pm$ 0.59 		& 79.80 $\pm$ 0.66 		& 83.42 $\pm$ 0.46 \\ 
CiteSeer 		& 54.41 $\pm$ 1.74 		& 54.87 $\pm$ 1.53 		& 54.73 $\pm$ 1.39 \\ 
 \hline
\end{tabular}}
\end{table}

\end{document}